\documentclass[10pt,twocolumn,letterpaper]{article}

\usepackage{iccv}
\usepackage{times}
\usepackage{epsfig}
\usepackage{graphicx}
\usepackage{amsmath}
\usepackage{amssymb}

\usepackage{booktabs}
\usepackage{subcaption}
\usepackage{adjustbox}
\usepackage{multirow}
\usepackage{bbding}
\usepackage[pagebackref,breaklinks,colorlinks]{hyperref}
\usepackage{enumitem}
\usepackage[linewidth=1pt]{mdframed}
\usepackage{float}
\usepackage[capitalize]{cleveref}
\crefname{section}{Sec.}{Secs.}
\Crefname{section}{Section}{Sections}
\Crefname{table}{Table}{Tables}
\crefname{table}{Tab.}{Tabs.}

\newcommand{\jiaze}{\textcolor{blue}}

\newcommand{\red}{\textcolor{blue}}

\iccvfinalcopy 

\def\iccvPaperID{5154} 

\ificcvfinal\fi

\begin{document}

\title{Traj-MAE: Masked Autoencoders for Trajectory Prediction}

\author{
Hao Chen$^{1*}$, Jiaze Wang$^{1*}$, Kun Shao$^{3}$, Furui Liu$^{2}$, Jianye Hao$^{3}$, \\ Chenyong Guan$^{4}$, Guangyong Chen$^{2\dag}$, Pheng-Ann Heng$^{1}$
}

\maketitle

\def\thefootnote{*}\footnotetext{These authors contributed equally to this work}\def\thefootnote{\arabic{footnote}}
\def\thefootnote{$\dag$}\footnotetext{Corresponding Author: gychen@zhejianglab.com}\def\thefootnote{\arabic{footnote}}
\def\thefootnote{1}\footnotetext{Computer Science and Engineering, The Chinese University of Hong Kong}\def\thefootnote{\arabic{footnote}}
\def\thefootnote{2}\footnotetext{Zhejiang Lab, Hangzhou, China}\def\thefootnote{\arabic{footnote}}
\def\thefootnote{3}\footnotetext{Huawei Noah’s Ark Lab}\def\thefootnote{\arabic{footnote}}
\def\thefootnote{4}\footnotetext{Gudsen Technology Co. Ltd}\def\thefootnote{\arabic{footnote}}
\ificcvfinal\thispagestyle{empty}\fi

\begin{abstract}
Trajectory prediction has been a crucial task in building a reliable autonomous driving system by anticipating possible dangers. One key issue is to generate consistent trajectory predictions without colliding. To overcome the challenge, we propose an efficient masked autoencoder for trajectory prediction (Traj-MAE) that better represents the complicated behaviors of agents in the driving environment. Specifically, our Traj-MAE employs diverse masking strategies to pre-train the trajectory encoder and map encoder, allowing for the capture of social and temporal information among agents while leveraging the effect of environment from multiple granularities. To address the catastrophic forgetting problem that arises when pre-training the network with multiple masking strategies, we introduce a continual pre-training framework, which can help Traj-MAE learn valuable and diverse information from various strategies efficiently. Our experimental results in both multi-agent and single-agent settings demonstrate that Traj-MAE achieves competitive results with state-of-the-art methods and significantly outperforms our baseline model. The code will be made publicly available upon publication.

\end{abstract}

\section{Introduction}
The goal of trajectory prediction is to predict the future trajectories of moving agents (e.g., \emph{pedestrians and
vehicles}), which is a crucial problem for building a safe, comfortable, and reliable autonomous driving system~\cite{liu2021multimodal,yuan2021agentformer,ngiam2021scene,choi2022hierarchical,sun2022m2i}. 
Many promising works~\cite{gu2021densetnt,cao2022advdo, shi2022motion, jia2021ide, zhang2022trajectory, wang2022ganet} have been proposed with great interest and demand from academia and industry.
It has been demonstrated that modeling complex interactions between agents \cite{salzmann2020trajectron++,saadatnejad2022socially,sadeghian2019sophie,bhattacharyya2021euro,jia2022multi} is of great importance in trajectory prediction. 
On this basis, to address the colliding prediction problem and generate consistent trajectory predictions, it is essential to model social and temporal relations between agents and to have a global understanding of maps \cite{azevedo2022exploiting}. In this paper, we investigate this issue using self-supervised learning.

\begin{figure}
    \centering
        \includegraphics[width=\linewidth]{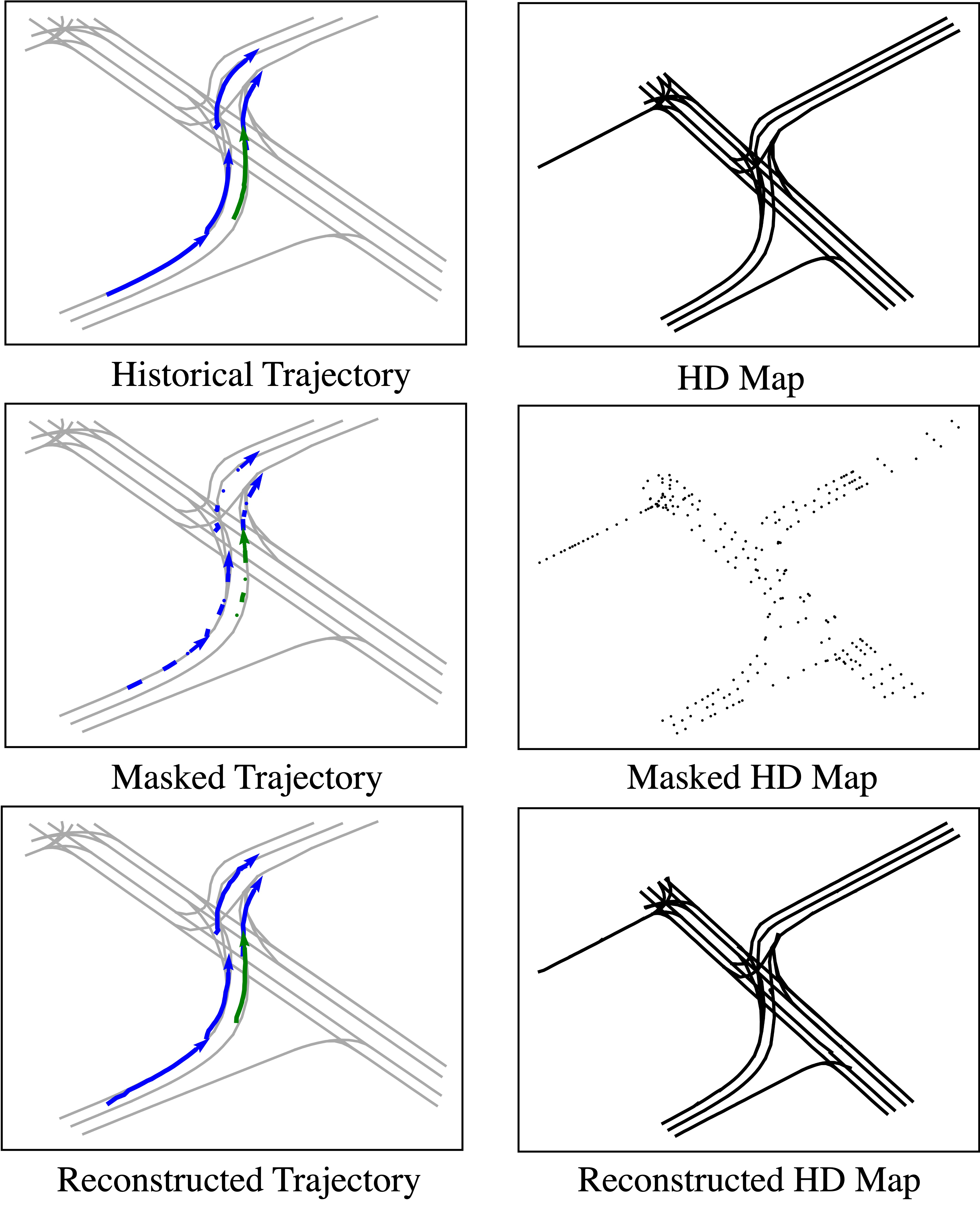}
    \caption{\textbf{An example of our masking and reconstruction strategy for trajectory prediction.} A portion of the historical trajectory and HD map is masked, the trajectory autoencoder and the map autoencoder are separately trained to recover the masked parts from the corrupted input. The green curve denotes the ego agent, the blue curves denote the surrounding agents, the same in the following figures in this paper.}
    \label{fig:teaser}
    \vspace{-12pt}
\end{figure}

Self-supervised learning aims to learn latent semantics from unlabeled data rather than building representations based on human annotations. Recent years have witnessed noteworthy advancements in the application of self-supervised learning to natural language processing \cite{devlin2018bert,yang2019xlnet} and computer vision \cite{wang2019learning,noroozi2016unsupervised,bao2021beit}.
One of the most promising self-supervised methods is the masked autoencoders~(MAE)~\cite{he2022masked} which achieve success in various tasks~\cite{pang2022masked,tong2022videomae}.
Furthermore, pre-train and finetune on the same small-scale datasets are also essential to learning a good representation \cite{el2021large}. 
Inspired by these works, we aim to explore the complex interactions between agents and the multiple granularities of maps using masked autoencoders. 
\emph{How to design an efficient masked autoencoder to generate consistent trajectory predictions?}
We attempt to answer the question from the following perspectives:

\textbf{(\romannumeral1)} The information density of trajectory and high definition (HD) maps differs significantly from that of images. While images are natural signals with high spatial redundancy, trajectories represent continuous temporal sequential signals with complex social interactions between agents, and HD maps contain highly structured information. Given the differences, models aimed at trajectory prediction require corresponding adjustments to capture informative features. Therefore, we investigate various masking strategies and suitable masking ratios for trajectories and HD maps. We develop both social and temporal masking to enable the trajectory encoder to capture information from diverse perspectives. We also study multiple granularities masking to enforce the map encoder to capture structural information from HD maps. Furthermore, we find that regardless of the masking strategy adopted, a high masking ratio (50\% $\sim$ 60\%) yields favorable results, which demands the encoders to acquire a holistic understanding of historical trajectories and HD maps.

\textbf{(\romannumeral2)} The absence of an efficient framework for pre-training multiple strategies poses a challenge for effective multimodal trajectory prediction. While traditional multi-task learning from scratch \cite{zhang2018overview} may struggle to converge due to the complex nature of this task, traditional continual learning methods \cite{chen2018lifelong,parisi2019continual} are limited by their inability to train the network with multiple tasks without forgetting previously learned knowledge. To address this issue, we propose a novel approach that trains the new strategy simultaneously with the original strategies, utilizing previously learned parameters to initialize the network. Therefore, we ensure that our network can acquire new knowledge while retaining previously obtained knowledge.

Driven by the analysis, we present \emph{Masked Trajectory Autoencoder~(Traj-MAE)}, an efficient and practical framework for self-supervised trajectory prediction.
As depicted in Figure \ref{fig:teaser}, Traj-MAE leverages partial masking of the input trajectory and HD map, utilizing the trajectory encoder and map encoder to reconstruct the masked segments, respectively. 
Through employing diverse masking strategies to reconstruct missing parts of the input trajectory and HD map, the trajectory encoder and map encoder can acquire a comprehensive understanding of the latent semantics of the inputs from various perspectives.
Moreover, we introduce a novel continual pre-training framework, which is a highly-efficient learning approach that trains the model with multiple strategies simultaneously, mitigating the issue of catastrophic forgetting.

Our core contributions are as follows:
\begin{itemize}[noitemsep,topsep=0pt]
\setlength{\itemsep}{0pt}
\item To our best knowledge, we are the first to present a neat and efficient masked trajectory autoencoder for self-supervised trajectory prediction.
\item We explore different masking strategies which fully utilize MAE to exploit the latent semantics of historical trajectory and HD map. Meanwhile, a continual pre-training framework is proposed to efficiently train the model with multiple strategies.
\item We conduct extensive experiments on the Argoverse, and INTERACTION for autonomous driving trajectory prediction, and the synthetic partition of TrajNet++ for pedestrian trajectory prediction. Our Traj-MAE achieves competitive results on these benchmarks and outperforms our baseline model by a notable margin. 
\end{itemize}


\begin{figure*}[t]
    \centering
    \includegraphics[width=1\linewidth]{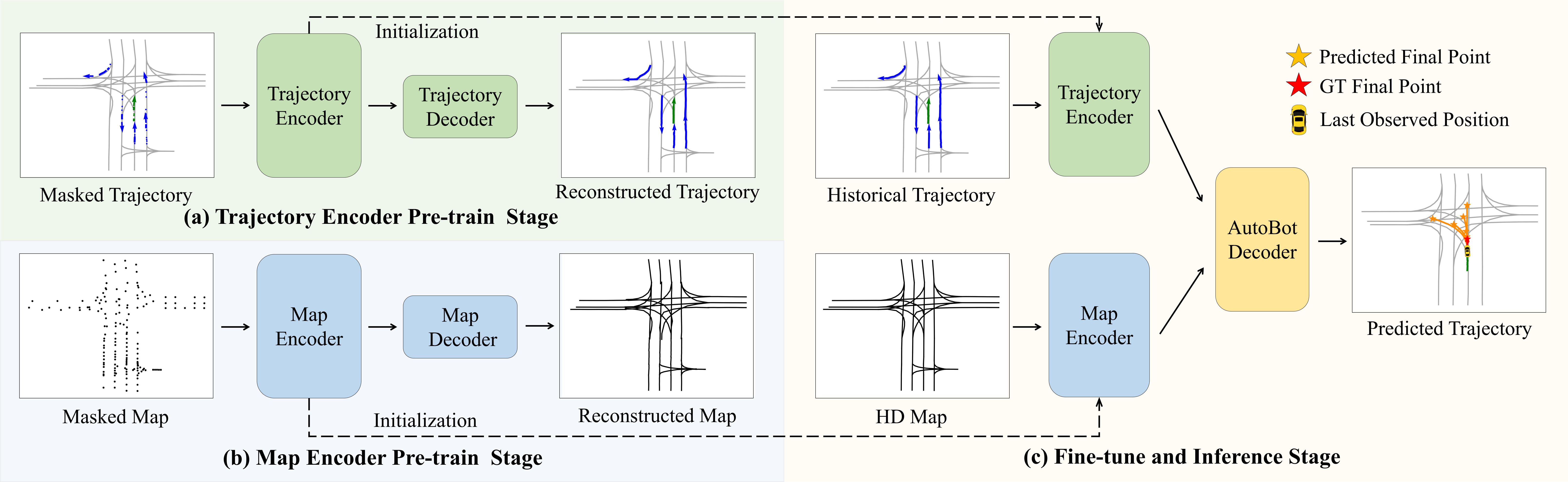}
    \caption{\textbf{Overview of Traj-MAE.} 
      Traj-MAE is mainly composed of three stages: (a) Trajectory encoder pre-train stage with continual trajectory masking and reconstruction strategies.  (b) Map encoder pre-train stage with continual map masking and reconstruction strategies. (c) Fine-tune and inference stage where the encoders are initialized by the pre-trained models' parameters.
    }
    \label{fig:framework}
    \vspace{-10pt}
\end{figure*}

\section{Related Works}

\noindent\textbf{Trajectory Prediction}
is widely considered a sequence modeling task with many RNN-based methods \cite{alahi2016social,zhang2019sr,messaoud2020multi,casas2020implicit} proposed to model the trajectory pattern of agents' future locations, as RNN (e.g., LSTM \cite{hochreiter1997long}) have achieved remarkable success in sequence modeling. 
Due to the strong ability of Transformers \cite{vaswani2017attention} to capture long-range dependencies, many transformer-based methods have emerged and flourished. STAR \cite{yu2020spatio} is proposed to capture complex spatio-temporal interactions by interleaving between spatial and temporal Transformers. mmTransformer \cite{liu2021multimodal} is designed to hierarchically aggregate the past trajectories, the road information, and the social interaction. For predicting multi-agents future trajectories, AgentFormer \cite{yuan2021agentformer} and AutoBots \cite{girgis2022latent} have given solutions to model the time dimension and social dimension simultaneously. The enhancement of the encoder's ability to model information in both dimensions is an interesting and central focus of this work.

\noindent\textbf{Self-supervised Learning}
has shown significant success in natural language processing and computer vision fields recently, especially the autoencoding method. Denoising autoencoders (DAE) \cite{vincent2008extracting} is a learning representation method that reconstructs original signals from corrupted inputs. BERT \cite{devlin2018bert} can be seen as a development of DAE, which masks input tokens and trains the model to predict the missing content. With the Masked Language Modeling (MLM) task proposed in BERT, many MLM variants \cite{yang2019xlnet,bao2020unilmv2} are proposed to improve the performance of transformer pre-training. Similarly, in computer vision, the autoencoding method often focuses on different pretext tasks for pre-training \cite{noroozi2016unsupervised,bao2021beit,he2022masked}. 
One of the most popular methods is MAE \cite{he2022masked}, which randomly masks input patches and trains the model to recover masked patches in pixel space. Continuous progress \cite{feichtenhofer2022masked,pang2022masked,bachmann2022multimae} based on MAE has verified its effectiveness. Following the concept of MAE, our approach concentrate on utilizing MAE as a tool to pre-train model encoders with powerful feature extraction capability.

\noindent\textbf{Continual Learning}
is a method to tackle the catastrophic forgetting problem that happens in sequentially learning samples of different input patterns. The methods can be roughly categorized into replay, regularization-based, and parameter isolation approaches \cite{de2021continual}. With respect to replay methods \cite{rebuffi2017icarl,rolnick2019experience,isele2018selective,chaudhry2019tiny,sun2020ernie}, previous task samples are replayed while learning a new task to alleviate forgetting. Instead, when learning new data, regularization-based methods \cite{silver2002task,rannen2017encoder,zeno2018task,kirkpatrick2017overcoming} often introduce a regularization term in the loss function to consolidate previous knowledge. Parameter isolation methods \cite{mallya2018packnet,serra2018overcoming} dedicate different model parameters to each task to prevent any possible forgetting. In this work, we propose a continual pre-training framework to tackle the forgetting problem, in which way we are able to improve the generalization of model encoders by leveraging the specific information contained in the training samples of related masking strategies.
\begin{figure*}[!ht]
    \centering
        \includegraphics[width=\linewidth]{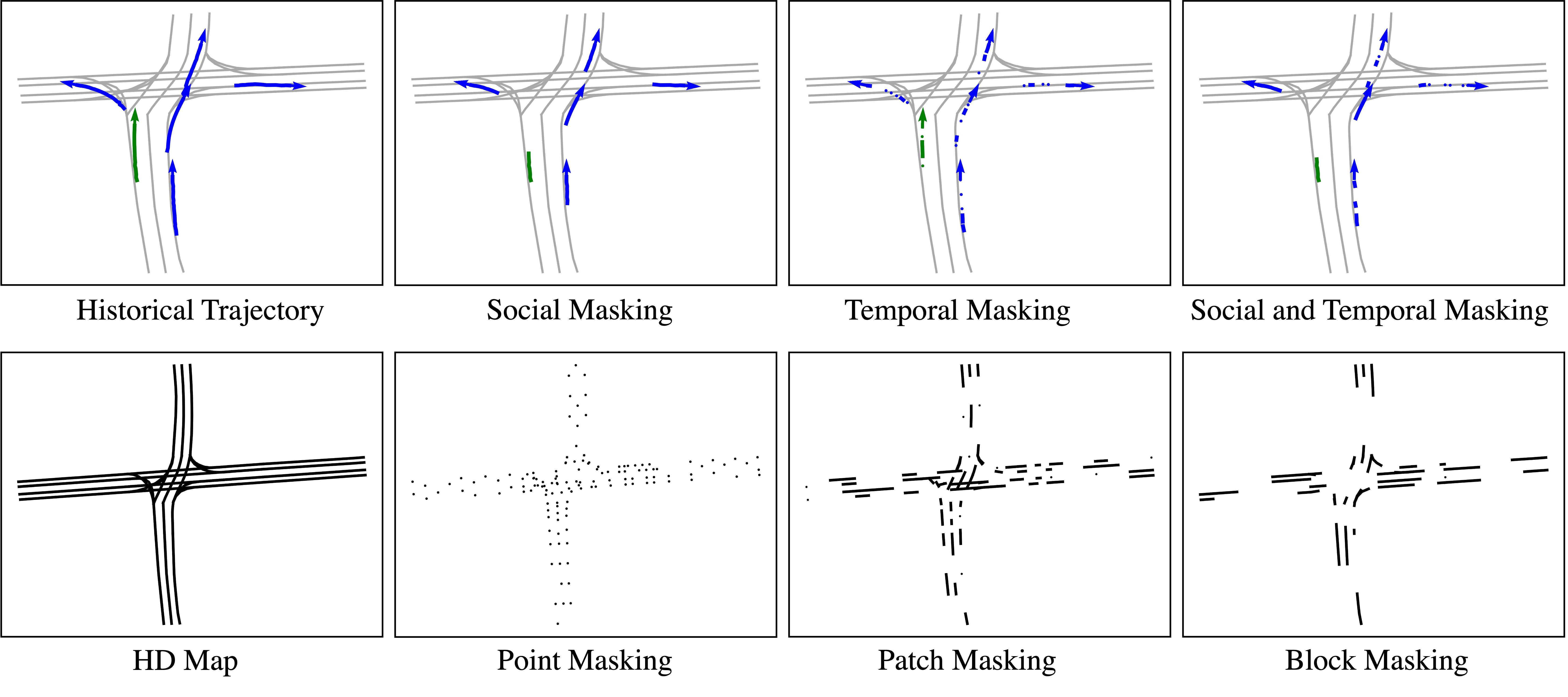}
    \caption{\textbf{Masking strategies for historical trajectory and HD map.} We show three masking strategies for the same input of trajectory and HD map, respectively. The leftmost column is input, the other three columns are different masking strategies performed on the input for historical trajectory (top) and HD map (bottom).}
    \label{fig:masking}
    \vspace{-10pt}
\end{figure*}

\section{Approach}
Our Traj-MAE is a sophisticated yet efficient self-supervised approach. Figure \ref{fig:framework} provides an overview of the Traj-MAE framework. In this section, we begin by introducing our network backbone. We then delve into our analysis of the masking strategies for trajectory and HD-map reconstruction. Finally, we discuss how we incorporate Traj-MAE into our continual pre-training framework.

\subsection{Network Backbone}
In this work, We use Autobots \cite{girgis2022latent} that has a transformer encoder-decoder architecture (detailed in supplementary material) as the baseline model to verify the effectiveness of the proposed method. 
Our Traj-MAE masks random parts from the input trajectory and HD map, then reconstructs the missing parts respectively.
Following MAE \cite{he2022masked} and VideoMAE~\cite{tong2022videomae}, we adopt the asymmetric encoder-decoder design to reduce computation. 

\noindent\textbf{Traj-MAE Encoder.} 
In Autobots, historical trajectories are encoded into context tensors, together with learnable seed parameters and map context, are passed to the decoder to predict future trajectories. Inspired by this design, we adopt the Autobots encoder as our trajectory encoder. However, in Autobots, the HD map is directly fed to the decoder, making it challenging for the model to capture the inherent information of the HD map. To address this limitation, we introduce a map encoder with a similar architecture to the trajectory encoder to better reconstruct the masked HD map. However, we observed that directly adding the map encoder to the Autobots results in little improvement (see supplementary material). Nevertheless, we found that pre-training the map encoder with our proposed masking and reconstruction strategy can further improve the accuracy, validating the effectiveness of our pre-training strategy.

\noindent\textbf{Traj-MAE Decoder.}
The encoder in Traj-MAE processes only the unmasked parts of the input, while the decoder reconstructs the missing parts from the latent representation and mask tokens. Mask tokens are shared vectors that indicate the presence of the missing parts that need to be predicted. Additionally, positional embeddings are added to all tokens to provide location information. Traj-MAE decoder is designed with Transformer blocks that are shallower than the encoder and are used solely during pre-training to perform the trajectory and map reconstruction strategies. This enables the decoder architecture to be flexible and independent of the encoder architecture. Pre-training with a lightweight decoder can notably reduce pre-training time.

\subsection{Masking Strategy.}
Different masking strategies determine the pretext task with different latent information that the network encoder can learn.
To capture the social and temporal information from historical trajectories and multiple granularity representations from HD maps,
we devise three masking strategies for the trajectory encoder and map encoder, respectively. 
Each strategy masks different scales and components of the input, with the goal of reconstructing the missing parts of the input.

\noindent\textbf{Trajectory Masking Strategy.}
We introduce three effective masking and reconstruction strategies to enhance representation learned by the trajectory encoder.
The three different strategies are illustrated in Figure \ref{fig:masking}.

\emph{Social Masking.} 
Understanding the social relationships between agents is a fundamental concern when predicting trajectories.
Social masking aims to reconstruct each agent's trajectory from surrounding agents. 
We mask the ego-agent's trajectory in the last observed time and nearby agents' trajectories at the beginning of the observed time.
By utilizing this strategy, the network is able to leverage continuous trajectories that are observable to reconstruct trajectories that are unobservable for other agents. This approach improves the network's ability to model interactions among agents and generate socially consistent predictions.

\emph{Temporal Masking.}
Temporal masking strategy endeavors to reconstruct the trajectories of all agents that have been randomly masked in the time domain. By inferring the positions of agents at various temporal intervals based on their positions at specific times, our model is able to efficiently capture the temporal dynamics of historical trajectories.

\emph{Social and Temporal Masking.} We also construct a masking strategy that reconstructs the historical trajectories in both temporal and social aspects. Specifically, half of the trajectories of surrounding agents are randomly masked in the time dimension, and half of that are masked in the last observed time, while the trajectory of the ego-agent in the last observed time is masked.
By reconstructing these trajectories simultaneously, the social and temporal masking can further enhance the trajectory encoder to obtain the temporal and social information. 

\noindent\textbf{Map Masking Strategy.}
The input vector of the map encoder is based on the VectorNet approach \cite{gao2020vectornet}, which selects a starting point and direction, and uniformly samples key points from the splines at the same spatial distance. To capture the latent semantics of the HD map, we propose a mask and reconstruction strategy that operates at multiple granularities.

\emph{Point Masking.} 
Our point masking strategy randomly samples and masks key points from the input map vector and reconstructs the whole map by predicting the missing points. This fine-grained learning approach is shown to have the best effect in our experiments.

\emph{Patch Masking.}
Patch masking refers to masking the map vector at the patch level, where patches are randomly sampled from the map vector. Patch masking and reconstruction are more complex than point masking due to the unknown surrounding points of the middle point. Consequently, the map encoder must infer the map architecture from longer distance points, allowing it to learn the long-term distance relation of the map.

\emph{Block Masking.} 
The block masking approach removes large blocks from the input and predicts the missing parts in the input HD map, enabling the map encoder to have a better global understanding of the whole map. Block masking and patch masking differ in their granularities. For each polyline in HD map, patch masking masking several consecutive line segments, whereas block masking only masks a single continuous line segment.



\noindent\textbf{Reconstruction Targets.}
Traj-MAE reconstructs the input by predicting the coordinates of location points in every masked part.
Our loss function comprises the Huber loss between the reconstructed trajectory and original trajectory to pre-train the trajectory encoder, and the Huber loss between the reconstructed HD map and original HD map to pre-train the map encoder.
Similar to MAE \cite{he2022masked} and Point-MAE \cite{pang2022masked}, we compute the loss only on masked parts.

\begin{figure}
    \centering
    \includegraphics[width=\linewidth]{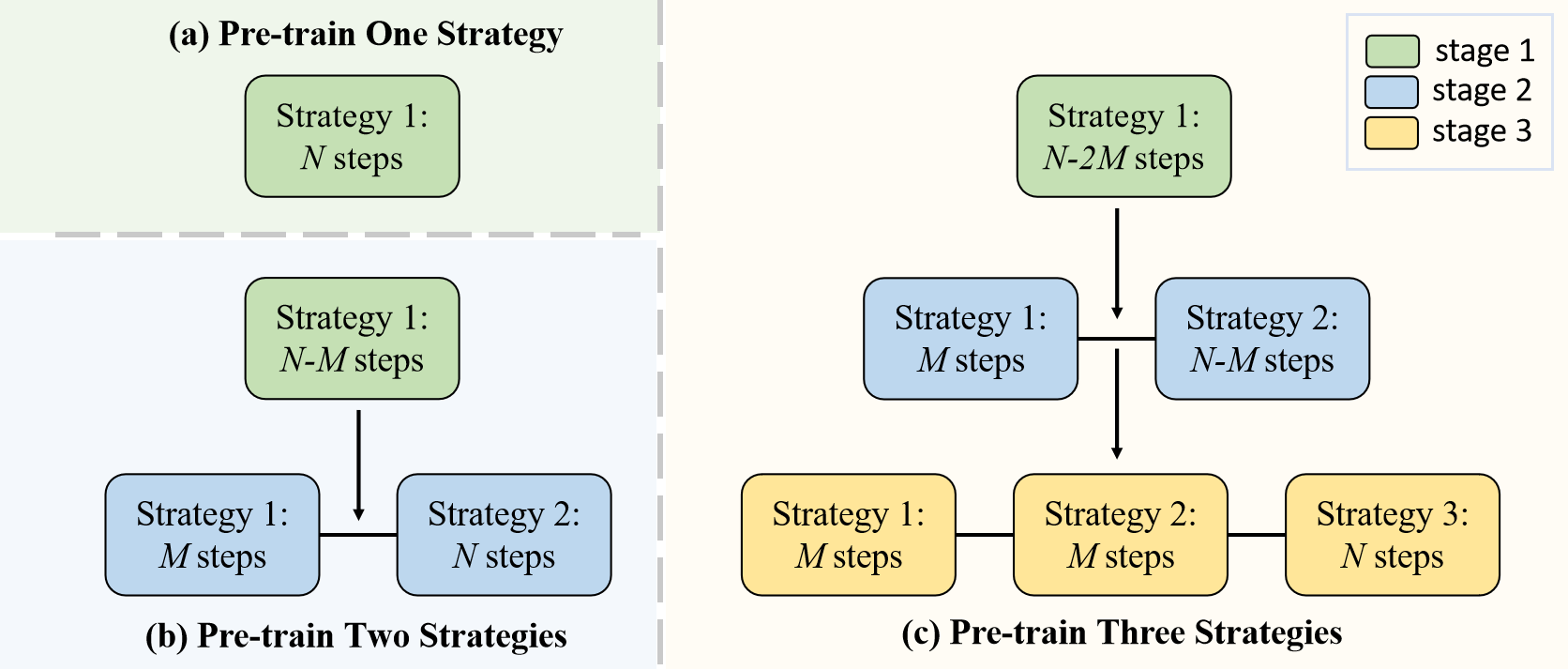}
    \caption{\textbf{Illustration of our continual pre-training framework.} For each strategy, we guarantee it has the same training steps throughout the pre-training process.}
    \label{fig:continue}
    \vspace{-10pt}
\end{figure}

\noindent\subsection{Continual Pre-training Framework} 
We propose a continual pre-training framework to learn diverse information from multiple masking strategies efficiently. 

Traditional continual learning method~\cite{chen2018lifelong,parisi2019continual} trains the model on only one strategy at each stage, and the model may suffer from the catastrophic forgetting problem, forgetting the previously learned knowledge.
Multi-task learning method \cite{zhang2018overview} trains the model with multiple tasks at the same time. In this way, we find it hard for our model to converge after many steps, and the performance could be even worse than the model pre-trained on a single strategy. 

Therefore, there are two issues to solve.
The first issue is to learn the strategies without forgetting the previous knowledge.
The second issue is how to pre-train the model with multiple strategies efficiently.
To overcome these issues, we propose a practical continual pre-training framework that enables model training with satisfactory efficiency and alleviates catastrophic forgetting. 
The core idea is to train the model over multiple stages with cross-stage parameter sharing. 
That means, for each pre-training stage except the first one, we use the parameters learned in previous stage to initialize the model. Then we pre-train a new strategy together with the previous strategies simultaneously in this stage. As shown in Figure \ref{fig:continue}, the number of pre-training stages is equal to that of strategies. Our framework allocates each strategy a fixed number of pre-training steps ($N$) and distributes the steps for each strategy over the whole pre-training stages. $M$ is the steps to pre-train the previous strategy in the new stage. 
To pre-train different strategies in a single stage, we randomly select training samples from each strategy. In this way, the effectiveness of our continual pre-training framework can be guaranteed.
\section{Experiments}
\subsection{Experimental Setup.}

\begin{table}[t]
\centering
\resizebox{\linewidth}{!}{%
\begin{tabular}{cccc} \toprule
Method & \hspace{-0.8cm}minADE & \hspace{-0.9cm}minFDE & \hspace{-0.9cm}MR \\ \hline
TNT \cite{zhao2020tnt} & \hspace{-0.8cm}0.94 & \hspace{-0.9cm}1.54 & \hspace{-0.9cm}0.13 \\
LaneRCNN \cite{zeng2021lanercnn} & \hspace{-0.8cm}0.90 & \hspace{-0.9cm}1.45 & \hspace{-0.9cm}0.12 \\
mmTransformer \cite{liu2021multimodal} &\hspace{-0.8cm} 0.84 & \hspace{-0.9cm}1.34 & \hspace{-0.9cm}0.15 \\
GOHOME \cite{gilles2022gohome} & \hspace{-0.8cm}0.94 & \hspace{-0.9cm}1.45 & \hspace{-0.9cm}\textbf{0.11} \\
TPCN \cite{ye2021tpcn} & \hspace{-0.8cm}0.87 & \hspace{-0.9cm}1.38 & \hspace{-0.9cm}0.16 \\
Scene Transformer \cite{ngiam2021scene} & \hspace{-0.8cm}0.80 & \hspace{-0.9cm}1.23 & \hspace{-0.9cm}0.13 \\
MultiPath++ \cite{varadarajan2022multipath++} & \hspace{-0.8cm}0.79 & \hspace{-0.9cm}1.21 & \hspace{-0.9cm}0.13        \\
DenseTNT \cite{gu2021densetnt} & \hspace{-0.8cm}0.88 & \hspace{-0.9cm}1.28 & \hspace{-0.9cm}0.13 \\ 
HiVT \cite{zhou2022hivt} & \hspace{-0.8cm}\textbf{0.77} & \hspace{-0.9cm}1.17 & \hspace{-0.9cm}0.13 \\ 
Wayformer \cite{nayakanti2022wayformer} & \hspace{-0.8cm}\textbf{0.77} & \hspace{-0.9cm}1.16 & \hspace{-0.9cm}0.12 \\
DCMS \cite{ye2022dcms} & \hspace{-0.8cm}\textbf{0.77} & \hspace{-0.9cm}\textbf{1.14} & \hspace{-0.9cm}\textbf{0.11} \\
GANet \cite{wang2022ganet} & \hspace{-0.8cm}0.81 & \hspace{-0.9cm}1.16 & \hspace{-0.9cm}0.12 \\ 
DSP \cite{zhang2022trajectory} & \hspace{-0.8cm}0.82 & \hspace{-0.9cm}1.22 & \hspace{-0.9cm}0.13 \\ \hline
Autobot-Ego \cite{girgis2022latent} & \hspace{-0.8cm}0.89 & \hspace{-0.9cm}1.41 & \hspace{-0.9cm}0.16 \\
\textbf{Traj-MAE} & \hspace{0.05cm}0.81 \red{$\downarrow$ 9\%} & \hspace{0.1cm}1.25 \red{$\downarrow$ 11\%} & \hspace{0.1cm}0.137 \red{$\downarrow$ 14\%}\\ \toprule
\end{tabular}
}
\caption{\textbf{Comparison with state-of-the-art methods on the Argoverse test set}.}
\label{table:argo_test}
\vspace{-10pt}
\end{table}

\begin{figure*}[t]
    \centering
    \includegraphics[width=\linewidth]{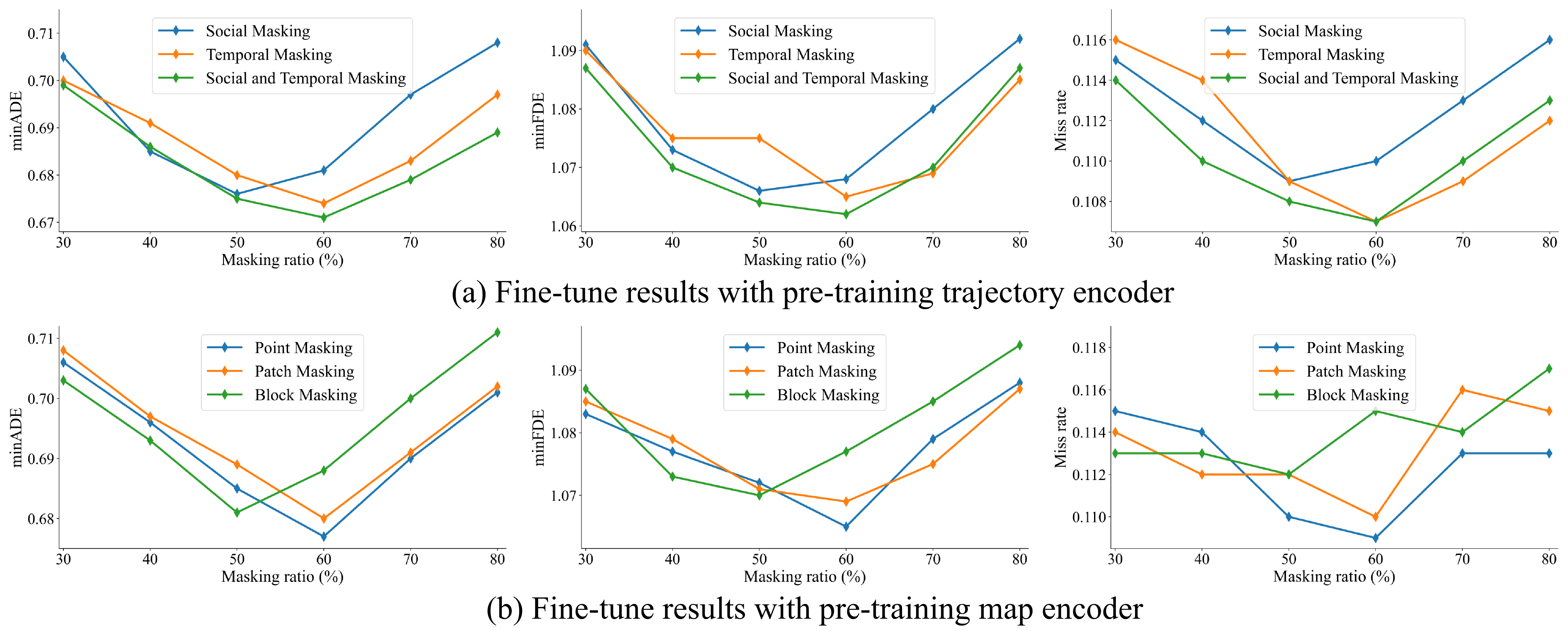}
    \caption{\textbf{Ablation study on masking strategy with different ratios.}}
    \label{fig:ablation}
    \vspace{3pt}
\end{figure*}

\begin{table*}[t]
\centering
\setlength{\tabcolsep}{0.01mm}{
\begin{tabular}{ccccccc} \toprule
Method & MinJointADE & MinJointFDE& MinJointMR & \begin{tabular}[c]{@{}c@{}}\hspace{-0.2cm}Cross\\ \hspace{-0.2cm}CollisionRate\end{tabular} & \begin{tabular}[c]{@{}c@{}}\hspace{-0.2cm}Ego\\ \hspace{-0.2cm}CollisionRate\end{tabular} & \begin{tabular}[c]{@{}c@{}}Consistent\\ MinJointMR\end{tabular} \\ \hline
DenseTNT \cite{gu2021densetnt} & 0.412 & 1.129 & 0.224 & \hspace{-0.2cm}\textbf{0} & \hspace{-0.2cm}0.014 & 0.224 \\
HDGT \cite{jia2022hdgt} & \textbf{0.303} & \textbf{0.958} & 0.194 & \hspace{-0.2cm}0.163 & \hspace{-0.2cm}0.005 & 0.236 \\
THOMAS \cite{gilles2021thomas} & 0.416 & 0.968 & \textbf{0.179} & \hspace{-0.2cm}0.128 & \hspace{-0.2cm}0.011 & 0.252 \\
ReCoG2 \cite{mo2020recog} & 0.467 & 1.160 & 0.238 & \hspace{-0.2cm}0.069 & \hspace{-0.2cm}0.011 & 0.268 \\
L-GCN & 0.393 & 1.249 & 0.284 & \hspace{-0.2cm}0.060 & \hspace{-0.2cm}\textbf{0.004} & 0.297 \\
MoliNet & 0.729 & 2.554 & 0.444 & \hspace{-0.2cm}0.075 & \hspace{-0.2cm}0.042 & 0.473 \\ 
HGT-Joint & 0.307 & 1.056 & 0.186 & \hspace{-0.2cm}0.016 & \hspace{-0.2cm}0.005 & 0.190 \\\hline
AutoBot \cite{girgis2022latent} & 0.312 & 1.015 & 0.193 & \hspace{-0.2cm}0.043 & \hspace{-0.2cm}0.010 & 0.207 \\
\textbf{Traj-MAE} & \hspace{0.82cm}0.306 \red{$\downarrow$ 2\%} & \hspace{0.82cm}0.966 \red{$\downarrow$ 5\%} & \hspace{0.82cm}0.183 \red{$\downarrow$ 5\%} & \hspace{0.82cm}0.021 \red{$\downarrow$ 51\%}  & \hspace{0.82cm}0.006 \red{$\downarrow$ 40\%} & \hspace{0.82cm}\textbf{0.188} \red{$\downarrow$ 9\%} \\ \toprule
\end{tabular}
}
\caption{\textbf{Comparison with state-of-the-art methods on the INTERACTION Multi-Agent Track}.}
\label{table:INTERACTION}
\vspace{-10pt}
\end{table*}

\begin{table}[t]
\centering
\resizebox{\linewidth}{!}{%
\begin{tabular}{cccc}
\toprule
Model & \begin{tabular}[c]{@{}c@{}}\hspace{-0.8cm}Ego \\ \hspace{-0.8cm}minADE\end{tabular} & \begin{tabular}[c]{@{}c@{}}\hspace{-1cm}Scene \\ \hspace{-1cm}minADE\end{tabular} & \begin{tabular}[c]{@{}c@{}}\hspace{-1cm}Scene \\\hspace{-1cm} minFDE\end{tabular} \\ \hline
Linear \cite{girgis2022latent} & \hspace{-1cm}0.439 & \hspace{-1cm}0.409 & \hspace{-1cm}0.897 \\
AutoBot-AS \cite{girgis2022latent} & \hspace{-1cm}0.196  & \hspace{-1cm}0.316 & \hspace{-1cm}0.632 \\
AutoBot-Ego \cite{girgis2022latent} & \hspace{-1cm}0.098 & \hspace{-1cm}0.214 & \hspace{-1cm}0.431 \\ \hline
AutoBot \cite{girgis2022latent} & \hspace{-1cm}0.095  & \hspace{-1cm}0.128 & \hspace{-1cm}0.234 \\
\textbf{Traj-MAE} & \textbf{0.074} \red{$\downarrow$ 22\%}  & \textbf{0.093} \red{$\downarrow$ 27\%} & \textbf{0.181} \red{$\downarrow$ 23\%} \\ \toprule
\end{tabular}
}
\caption{\textbf{Results on TrajNet++ for a multi-agent forecasting scenario}.}
\label{table:Traj++}
\vspace{-10pt}
\end{table}

\noindent\textbf{Datasets.} Argoverse motion forecasting dataset \cite{chang2019argoverse} provides 333K real-world driving sequences, which are sampled at 10Hz, with 2 seconds history and 3 seconds future. The whole dataset is split into train, validation and test sets, with 205942, 39472, and 78143 sequences, respectively.
The Interaction dataset \cite{zhan2019interaction} consists of various highly interactive driving situations, and each trajectory has 1 second history and 3 seconds future sampled at 10Hz. In the multi-agent prediction track, the target is to predict multiple target agents' coordinates and yaw jointly.
The synthetic partition of the TrajNet++ dataset \cite{sadeghian2018trajnet} has 54513 scenes, which is specifically designed to have a high level of interactions \cite{kothari2021human}. Given the state of all agents of the past 9 timesteps, the goal is to predict the next 12 timesteps for all agents. Due to TrajNet++ does no have HD map, we pre-train and fine-tune the trajectory encoder only.

\subsection{Experimental Results.}~\label{section:results}
In this subsection, we pre-train and fine-tune the model on the same benchmark to validate the benefit brought by our proposed self-supervised learning method. We forecast 6 future trajectories on all datasets. The meaning of the used metrics is presented in the supplementary material and the lower metrics indicate better performance.

\noindent\textbf{Results on Argoverse.}
The trajectory prediction results on Argoverse are reported in Table \ref{table:argo_test}. 
On the Argoverse test set, our Traj-MAE decreases the minADE by 9\%, minFDE by 11\% and MR by 14\%, respectively.
Although Traj-MAE does not achieve state-of-the-art, the performance demonstrates that our Traj-MAE significantly improves our baseline model (Autobot-Ego). Moreover, it uses much less computation, which is an effective and GPU-friendly pre-training mechanism that only pre-trained on a single GPU (V100) in under \emph{48h}. Additionally, the performance improvement achieved by Traj-MAE is also demonstrated on the validation set, as shown in \ref{ablation}.

\begin{figure*}[t]
    \centering
        \includegraphics[width=\linewidth]{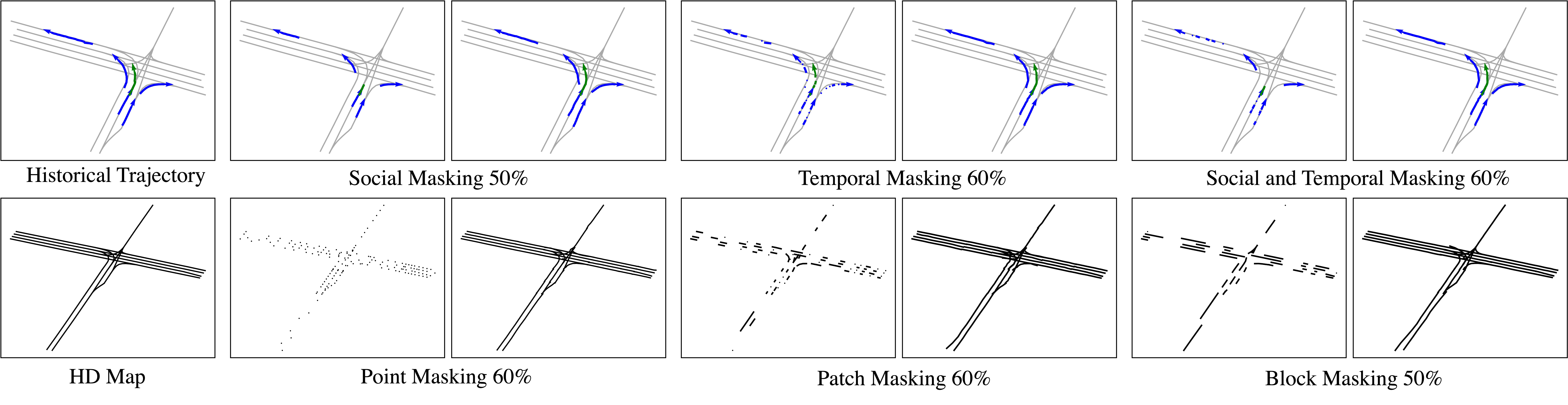}
    \caption{\textbf{Reconstruction results on historical trajectory and HD map with different masking strategies.} The leftmost side is input, and  the masked input (left) and reconstruction (right) with different masking strategies are shown in the following columns.}
    \label{fig:vis_strategy}
    \vspace{-1pt}
\end{figure*}

\begin{table*}[!]
\centering
\begin{subtable}[c]{0.48\textwidth}
\centering
\begin{tabular}{cccc}
\begin{tabular}{cccc}
\hline
      & minADE & minFDE & MR    \\ \hline
S    & 0.676   & 1.066   & 0.109 \\
T    & 0.674   & 1.065   & 0.107 \\
ST    & 0.671  & 1.062   & 0.107 \\
S $\to$ T & 0.642   & 1.042  & 0.103\\
T $\to$ S  & 0.651  & 1.051  &  0.105\\
S $\to$ T$\to$ ST & \textbf{0.621}   & \textbf{1.027}  & \textbf{0.099} \\
ST $\to$ T $\to$ S & 0.636   & 1.038   & 0.101 \\ \hline
\end{tabular}
\end{tabular}
\subcaption{Continual pre-training for \textbf{trajectory reconstruction}.}
\end{subtable}
\begin{subtable}[c]{0.48\textwidth}
\centering
\begin{tabular}{cccc}
\hline
      & minADE & minFDE & MR    \\ \hline
Po    & 0.677   & 1.065   & 0.109 \\
Pa    & 0.680   & 1.069   & 0.110 \\
B    & 0.681   & 1.070   & 0.112 \\
Po $\to$ Pa & 0.649   & 1.047   & 0.106 \\
Pa $\to$ Po &  0.651  & 1.053  &  0.108 \\
Po $\to$ Pa $\to$ B & \textbf{0.627}   & \textbf{1.033}   & \textbf{0.102} \\
B $\to$ Pa $\to$ Po & 0.641   & 1.046   & 0.104 \\ \hline
\end{tabular}
\subcaption{Continual pre-training for \textbf{HD map reconstruction}.}
\end{subtable}

\caption{\textbf{Ablation study on continual pre-training framework}. We show the results using different pre-train strategies with their best masking ratio. Note that 'S', 'T', 'ST' represent social masking, temporal masking, social and temporal masking strategy, respectively. 'Po', 'Pa', 'B' represent point masking, patch masking, block masking strategy, respectively.}
\label{Table:continue}
\vspace{-10pt}
\end{table*}
\noindent\textbf{Results on INTERACTION.}
In Table \ref{table:INTERACTION},
we evaluate our method on the INTERACTION multi-agent track and achieve state-of-the-art performance on this benchmark.
The table shows that Traj-MAE outperforms all other approaches in terms of Consistent MinJointMR, the ranking metric.
This metric encourages the model to make consistent predictions, thus the best result on this metric indicates that our Traj-MAE can well capture multi-agents' interaction.
As for other metrics, our model still has competitive results.
Especially, the reductions of 51\% and 40\% in CrossCollisionRate and EgoCollisionRate demonstrate that our Traj-MAE can better utilize social information and avoid collisions between agents.

\noindent\textbf{Results on TrajNet++.}
To further demonstrate that our Traj-MAE has the ability to capture the social and temporal information for multi-agent trajectory prediction.
We evaluate our method on the synthetic partition of TrajNet++ dataset.
We compare our model with linear extrapolation (Linear), our baseline Autobot,  and its variants: AutoBot-AntiSocial (AutoBot-AS) and AutoBot-Ego.
The experiment results of agent-level metric and scene-level metrics defined by \cite{casas2020implicit} are reported in Table \ref{table:Traj++}.
Our Traj-MAE achieves large reductions of 22\%, 27\%, 23\% with respect to Ego Agent's minADE, Scene-level minADE, and Scene-level minFDE respectively.

\subsection{Ablation Studies.}
\label{ablation}
To investigate the properties of our method, we perform in-depth ablation studies on the Argoverse validation set. 
For these experiments, all models share the same experiment settings and architecture.
We evaluate our models on minADE, minFDE, and miss rate (MR) of the predicted $6$ trajectories.
Our baseline model (Autobot-Ego with Map Encoder) achieves \textbf{0.732}, \textbf{1.096} and \textbf{0.119} on \textbf{minADE}, \textbf{minFDE} and \textbf{MR}, respectively.
When fine-tuning the trajectory encoder with the pre-trained model, the map encoder is trained from scratch and vice versa.

\noindent\textbf{Trajectory Masking Strategy.}
In Figure \ref{fig:ablation}(a), we compare the performance of different trajectory masking strategies and ratios. As the masking ratio increases from 30\% to a threshold of 50-60\%, the performance of all three strategies improves simultaneously. However, the performance degrades as the masking ratio increases beyond the threshold to 80\%. Furthermore, our experiments show that social and temporal masking outperforms the other two strategies, particularly in terms of minADE and minFDE. This suggests that incorporating social and temporal information is crucial for effective trajectory prediction. We attribute this to the fact that social and temporal masking encourages the trajectory encoder to capture more essential temporal and social information, thereby making Traj-MAE a challenging self-supervised trajectory prediction framework.

\begin{figure*}[!ht]
    \centering
        \includegraphics[width=\linewidth]{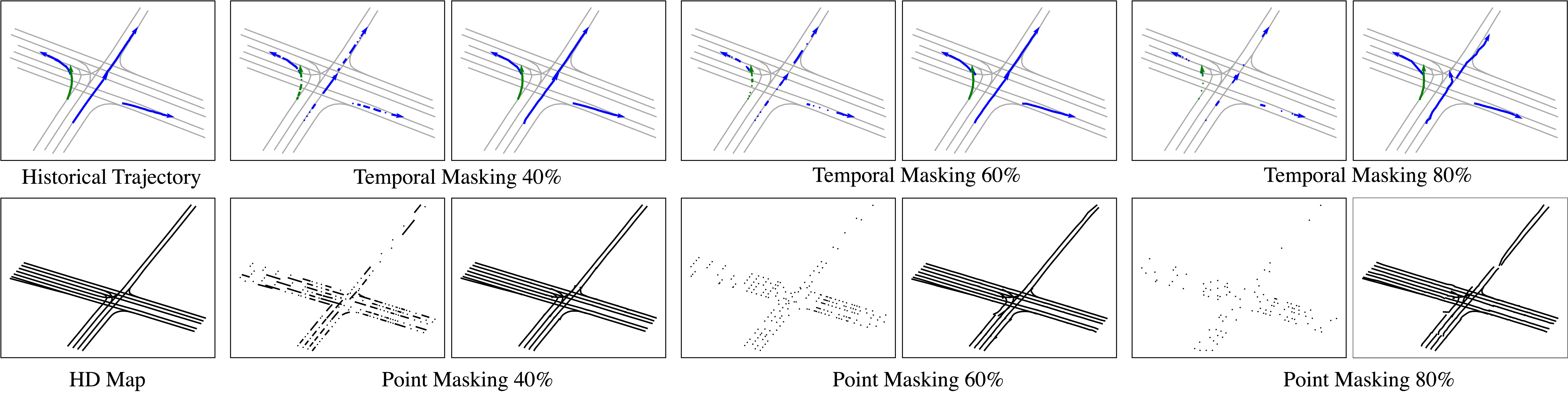}
    \caption{\textbf{Reconstruction results on historical trajectory and HD map with different masking ratios.} The leftmost side is input, and  the masked input (left) and reconstruction (right) with different masking ratios are shown in the following columns.}
    \label{fig:vis_ratio}
    \vspace{-10pt}
\end{figure*}

\begin{figure}[!ht]
    \vspace{5pt}
    \centering
        \includegraphics[width=\linewidth]{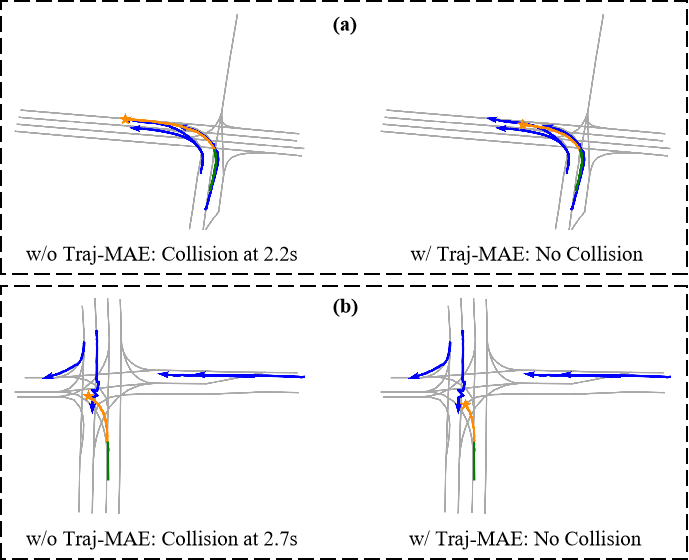}
    \caption{\textbf{Examples of collision avoidance using Traj-MAE.} The result in the left side is predicted by our baseline model and the right side is the future motion predicted using Traj-MAE. The predicted trajectory is denoted by orange curve.}
    \label{fig:collision}
    \vspace{-15pt}
\end{figure}

\noindent\textbf{Map Masking Strategy.}
Figure \ref{fig:ablation}(b) shows the influence of masking strategies with different masking ratios for HD map.
Point masking works best for pre-training our map encoder.
Besides, point masking and patch masking allow for a higher masking ratio (60\%) compared to block masking (50\%), which can provide a more significant speedup benefit. We suppose that with the increase of connected masked parts, the mask and reconstruct task is more challenging, and the optimal mask ratio is smaller. 


\noindent\textbf{Continual Pre-training Strategy.}
Table \ref{Table:continue} shows the effectiveness of our Traj-MAE.
First, we find that our continual pre-training framework further improves the network performance than pre-training on a single strategy.
With continual pre-training, the trajectory encoder is better equipped to capture social and temporal information, while the map encoder excels at capturing structured information from multiple granularities.
Moreover, our experiments reveal that the sequence of pre-training strategies can also impact model performance.
For trajectory encoder pre-training, the best sequence is \emph{social masking $\to$ temporal masking $\to$ social and temporal masking}.
For map encoder pre-training, the best sequence is 
\emph{point masking $\to$ patch masking $\to$ block masking}.
By integrating our pre-trained trajectory and map encoders, we achieve impressive results on the Argoverse validation set, with a \textbf{minADE} of \textbf{0.604}, a \textbf{minFDE} of \textbf{1.003}, and an \textbf{MR} of \textbf{0.092}.

\noindent\textbf{Qualitative Analysis.}
We visualize the reconstruction results on the Argoverse validation set for qualitative analysis, where Figure \ref{fig:vis_strategy} illustrates the different masking strategies.
Although social masking is challenging for trajectory reconstruction and block masking is difficult for map reconstruction, adopting a moderate masking ratio with these strategies can still yield satisfactory reconstruction results.
We compare the different temporal masking ratios and point masking ratios in Figure \ref{fig:vis_ratio}.
Our Traj-MAE can produce satisfying reconstructed trajectories and HD maps even under a high masking ratio (e.g., 80\%), which indicates that our Traj-MAE can learn useful high-level representations.
Furthermore, in Figure \ref{fig:collision}, we present two examples to demonstrate how Traj-MAE can effectively reduce the occurrence of collisions. Specifically, in a scene where our baseline model predicts a collision, Traj-MAE leverages the interaction relations among agents' historical trajectories to avoid collision happening.

\section{Conclusion}
This paper proposes a novel masked trajectory autoencoder (Traj-MAE) for self-supervised trajectory prediction learning.
Our Traj-MAE incorporates diverse masking strategies that facilitate the trajectory encoder learning the social and temporal information and map encoder capturing structural information with multiple granularities. We also propose a continual pre-training framework that enables efficient pre-training of multiple strategies. Experimental results show that our Traj-MAE produces impressive results on various challenging datasets in both multiple-agent and single-agent settings. We hope that this work will inspire further investigation into self-supervised learning for trajectory prediction.




\section*{Acknowledgement}
The work was supported by National Key R\&D Program of China (2022YFB4501500, 2022YFB4501504)

{\small
\bibliographystyle{ieee_fullname}
\bibliography{egbib}

\begin{thebibliography}{10}\itemsep=-1pt

\bibitem{alahi2016social}
Alexandre Alahi, Kratarth Goel, Vignesh Ramanathan, Alexandre Robicquet, Li
  Fei-Fei, and Silvio Savarese.
\newblock Social lstm: Human trajectory prediction in crowded spaces.
\newblock In {\em Proceedings of the IEEE conference on computer vision and
  pattern recognition}, pages 961--971, 2016.

\bibitem{azevedo2022exploiting}
Caio Azevedo, Thomas Gilles, Stefano Sabatini, and Dzmitry Tsishkou.
\newblock Exploiting map information for self-supervised learning in motion
  forecasting.
\newblock {\em arXiv preprint arXiv:2210.04672}, 2022.

\bibitem{bachmann2022multimae}
Roman Bachmann, David Mizrahi, Andrei Atanov, and Amir Zamir.
\newblock Multimae: Multi-modal multi-task masked autoencoders.
\newblock {\em arXiv preprint arXiv:2204.01678}, 2022.

\bibitem{bao2021beit}
Hangbo Bao, Li Dong, and Furu Wei.
\newblock Beit: Bert pre-training of image transformers.
\newblock {\em arXiv preprint arXiv:2106.08254}, 2021.

\bibitem{bao2020unilmv2}
Hangbo Bao, Li Dong, Furu Wei, Wenhui Wang, Nan Yang, Xiaodong Liu, Yu Wang,
  Jianfeng Gao, Songhao Piao, Ming Zhou, et~al.
\newblock Unilmv2: Pseudo-masked language models for unified language model
  pre-training.
\newblock In {\em International Conference on Machine Learning}, pages
  642--652. PMLR, 2020.

\bibitem{bhattacharyya2021euro}
Apratim Bhattacharyya, Daniel~Olmeda Reino, Mario Fritz, and Bernt Schiele.
\newblock Euro-pvi: Pedestrian vehicle interactions in dense urban centers.
\newblock In {\em Proceedings of the IEEE/CVF Conference on Computer Vision and
  Pattern Recognition}, pages 6408--6417, 2021.

\bibitem{cao2022advdo}
Yulong Cao, Chaowei Xiao, Anima Anandkumar, Danfei Xu, and Marco Pavone.
\newblock Advdo: Realistic adversarial attacks for trajectory prediction.
\newblock In {\em European Conference on Computer Vision}, pages 36--52.
  Springer, 2022.

\bibitem{casas2020implicit}
Sergio Casas, Cole Gulino, Simon Suo, Katie Luo, Renjie Liao, and Raquel
  Urtasun.
\newblock Implicit latent variable model for scene-consistent motion
  forecasting.
\newblock In {\em European Conference on Computer Vision}, pages 624--641.
  Springer, 2020.

\bibitem{chang2019argoverse}
Ming-Fang Chang, John Lambert, Patsorn Sangkloy, Jagjeet Singh, Slawomir Bak,
  Andrew Hartnett, De Wang, Peter Carr, Simon Lucey, Deva Ramanan, et~al.
\newblock Argoverse: 3d tracking and forecasting with rich maps.
\newblock In {\em Proceedings of the IEEE/CVF Conference on Computer Vision and
  Pattern Recognition}, pages 8748--8757, 2019.

\bibitem{chaudhry2019tiny}
Arslan Chaudhry, Marcus Rohrbach, Mohamed Elhoseiny, Thalaiyasingam Ajanthan,
  Puneet~K Dokania, Philip~HS Torr, and Marc'Aurelio Ranzato.
\newblock On tiny episodic memories in continual learning.
\newblock {\em arXiv preprint arXiv:1902.10486}, 2019.

\bibitem{chen2018lifelong}
Zhiyuan Chen and Bing Liu.
\newblock Lifelong machine learning.
\newblock {\em Synthesis Lectures on Artificial Intelligence and Machine
  Learning}, 12(3):1--207, 2018.

\bibitem{choi2022hierarchical}
Dooseop Choi and KyoungWook Min.
\newblock Hierarchical latent structure for multi-modal vehicle trajectory
  forecasting.
\newblock In {\em European Conference on Computer Vision}, pages 129--145.
  Springer, 2022.

\bibitem{de2021continual}
Matthias De~Lange, Rahaf Aljundi, Marc Masana, Sarah Parisot, Xu Jia,
  Ale{\v{s}} Leonardis, Gregory Slabaugh, and Tinne Tuytelaars.
\newblock A continual learning survey: Defying forgetting in classification
  tasks.
\newblock {\em IEEE transactions on pattern analysis and machine intelligence},
  44(7):3366--3385, 2021.

\bibitem{devlin2018bert}
Jacob Devlin, Ming-Wei Chang, Kenton Lee, and Kristina Toutanova.
\newblock Bert: Pre-training of deep bidirectional transformers for language
  understanding.
\newblock {\em arXiv preprint arXiv:1810.04805}, 2018.

\bibitem{el2021large}
Alaaeldin El-Nouby, Gautier Izacard, Hugo Touvron, Ivan Laptev, Herv{\'e}
  Jegou, and Edouard Grave.
\newblock Are large-scale datasets necessary for self-supervised pre-training?
\newblock {\em arXiv preprint arXiv:2112.10740}, 2021.

\bibitem{feichtenhofer2022masked}
Christoph Feichtenhofer, Haoqi Fan, Yanghao Li, and Kaiming He.
\newblock Masked autoencoders as spatiotemporal learners.
\newblock {\em arXiv preprint arXiv:2205.09113}, 2022.

\bibitem{gao2020vectornet}
Jiyang Gao, Chen Sun, Hang Zhao, Yi Shen, Dragomir Anguelov, Congcong Li, and
  Cordelia Schmid.
\newblock Vectornet: Encoding hd maps and agent dynamics from vectorized
  representation.
\newblock In {\em Proceedings of the IEEE/CVF Conference on Computer Vision and
  Pattern Recognition}, pages 11525--11533, 2020.

\bibitem{gilles2021thomas}
Thomas Gilles, Stefano Sabatini, Dzmitry Tsishkou, Bogdan Stanciulescu, and
  Fabien Moutarde.
\newblock Thomas: Trajectory heatmap output with learned multi-agent sampling.
\newblock {\em arXiv preprint arXiv:2110.06607}, 2021.

\bibitem{gilles2022gohome}
Thomas Gilles, Stefano Sabatini, Dzmitry Tsishkou, Bogdan Stanciulescu, and
  Fabien Moutarde.
\newblock Gohome: Graph-oriented heatmap output for future motion estimation.
\newblock In {\em 2022 International Conference on Robotics and Automation
  (ICRA)}, pages 9107--9114. IEEE, 2022.

\bibitem{girgis2022latent}
Roger Girgis, Florian Golemo, Felipe Codevilla, Martin Weiss, Jim~Aldon
  D'Souza, Samira~Ebrahimi Kahou, Felix Heide, and Christopher Pal.
\newblock Latent variable sequential set transformers for joint multi-agent
  motion prediction.
\newblock In {\em International Conference on Learning Representations}, 2022.

\bibitem{gu2021densetnt}
Junru Gu, Chen Sun, and Hang Zhao.
\newblock Densetnt: End-to-end trajectory prediction from dense goal sets.
\newblock In {\em Proceedings of the IEEE/CVF International Conference on
  Computer Vision}, pages 15303--15312, 2021.

\bibitem{he2022masked}
Kaiming He, Xinlei Chen, Saining Xie, Yanghao Li, Piotr Doll{\'a}r, and Ross
  Girshick.
\newblock Masked autoencoders are scalable vision learners.
\newblock In {\em Proceedings of the IEEE/CVF Conference on Computer Vision and
  Pattern Recognition}, pages 16000--16009, 2022.

\bibitem{hochreiter1997long}
Sepp Hochreiter and J{\"u}rgen Schmidhuber.
\newblock Long short-term memory.
\newblock {\em Neural computation}, 9(8):1735--1780, 1997.

\bibitem{isele2018selective}
David Isele and Akansel Cosgun.
\newblock Selective experience replay for lifelong learning.
\newblock In {\em Proceedings of the AAAI Conference on Artificial
  Intelligence}, volume~32, 2018.

\bibitem{jia2021ide}
Xiaosong Jia, Liting Sun, Masayoshi Tomizuka, and Wei Zhan.
\newblock Ide-net: Interactive driving event and pattern extraction from human
  data.
\newblock {\em IEEE Robotics and Automation Letters}, 6(2):3065--3072, 2021.

\bibitem{jia2022multi}
Xiaosong Jia, Liting Sun, Hang Zhao, Masayoshi Tomizuka, and Wei Zhan.
\newblock Multi-agent trajectory prediction by combining egocentric and
  allocentric views.
\newblock In {\em Conference on Robot Learning}, pages 1434--1443. PMLR, 2022.

\bibitem{jia2022hdgt}
Xiaosong Jia, Penghao Wu, Li Chen, Hongyang Li, Yu Liu, and Junchi Yan.
\newblock Hdgt: Heterogeneous driving graph transformer for multi-agent
  trajectory prediction via scene encoding.
\newblock {\em arXiv preprint arXiv:2205.09753}, 2022.

\bibitem{kirkpatrick2017overcoming}
James Kirkpatrick, Razvan Pascanu, Neil Rabinowitz, Joel Veness, Guillaume
  Desjardins, Andrei~A Rusu, Kieran Milan, John Quan, Tiago Ramalho, Agnieszka
  Grabska-Barwinska, et~al.
\newblock Overcoming catastrophic forgetting in neural networks.
\newblock {\em Proceedings of the national academy of sciences},
  114(13):3521--3526, 2017.

\bibitem{kothari2021human}
Parth Kothari, Sven Kreiss, and Alexandre Alahi.
\newblock Human trajectory forecasting in crowds: A deep learning perspective.
\newblock {\em IEEE Transactions on Intelligent Transportation Systems},
  23(7):7386--7400, 2021.

\bibitem{liu2021multimodal}
Yicheng Liu, Jinghuai Zhang, Liangji Fang, Qinhong Jiang, and Bolei Zhou.
\newblock Multimodal motion prediction with stacked transformers.
\newblock In {\em Proceedings of the IEEE/CVF Conference on Computer Vision and
  Pattern Recognition}, pages 7577--7586, 2021.

\bibitem{mallya2018packnet}
Arun Mallya and Svetlana Lazebnik.
\newblock Packnet: Adding multiple tasks to a single network by iterative
  pruning.
\newblock In {\em Proceedings of the IEEE conference on Computer Vision and
  Pattern Recognition}, pages 7765--7773, 2018.

\bibitem{messaoud2020multi}
Kaouther Messaoud, Nachiket Deo, Mohan~M Trivedi, and Fawzi Nashashibi.
\newblock Multi-head attention with joint agent-map representation for
  trajectory prediction in autonomous driving.
\newblock {\em arXiv preprint arXiv:2005.02545}, 2020.

\bibitem{mo2020recog}
Xiaoyu Mo, Yang Xing, and Chen Lv.
\newblock Recog: A deep learning framework with heterogeneous graph for
  interaction-aware trajectory prediction.
\newblock {\em arXiv preprint arXiv:2012.05032}, 2020.

\bibitem{nayakanti2022wayformer}
Nigamaa Nayakanti, Rami Al-Rfou, Aurick Zhou, Kratarth Goel, Khaled~S Refaat,
  and Benjamin Sapp.
\newblock Wayformer: Motion forecasting via simple \& efficient attention
  networks.
\newblock {\em arXiv preprint arXiv:2207.05844}, 2022.

\bibitem{ngiam2021scene}
Jiquan Ngiam, Benjamin Caine, Vijay Vasudevan, Zhengdong Zhang, Hao-Tien~Lewis
  Chiang, Jeffrey Ling, Rebecca Roelofs, Alex Bewley, Chenxi Liu, Ashish
  Venugopal, et~al.
\newblock Scene transformer: A unified architecture for predicting multiple
  agent trajectories.
\newblock {\em arXiv preprint arXiv:2106.08417}, 2021.

\bibitem{noroozi2016unsupervised}
Mehdi Noroozi and Paolo Favaro.
\newblock Unsupervised learning of visual representations by solving jigsaw
  puzzles.
\newblock In {\em European conference on computer vision}, pages 69--84.
  Springer, 2016.

\bibitem{pang2022masked}
Yatian Pang, Wenxiao Wang, Francis~EH Tay, Wei Liu, Yonghong Tian, and Li Yuan.
\newblock Masked autoencoders for point cloud self-supervised learning.
\newblock {\em ECCV}, 2022.

\bibitem{parisi2019continual}
German~I Parisi, Ronald Kemker, Jose~L Part, Christopher Kanan, and Stefan
  Wermter.
\newblock Continual lifelong learning with neural networks: A review.
\newblock {\em Neural Networks}, 113:54--71, 2019.

\bibitem{rannen2017encoder}
Amal Rannen, Rahaf Aljundi, Matthew~B Blaschko, and Tinne Tuytelaars.
\newblock Encoder based lifelong learning.
\newblock In {\em Proceedings of the IEEE International Conference on Computer
  Vision}, pages 1320--1328, 2017.

\bibitem{rebuffi2017icarl}
Sylvestre-Alvise Rebuffi, Alexander Kolesnikov, Georg Sperl, and Christoph~H
  Lampert.
\newblock icarl: Incremental classifier and representation learning.
\newblock In {\em Proceedings of the IEEE conference on Computer Vision and
  Pattern Recognition}, pages 2001--2010, 2017.

\bibitem{rolnick2019experience}
David Rolnick, Arun Ahuja, Jonathan Schwarz, Timothy Lillicrap, and Gregory
  Wayne.
\newblock Experience replay for continual learning.
\newblock {\em Advances in Neural Information Processing Systems}, 32, 2019.

\bibitem{saadatnejad2022socially}
Saeed Saadatnejad, Mohammadhossein Bahari, Pedram Khorsandi, Mohammad Saneian,
  Seyed-Mohsen Moosavi-Dezfooli, and Alexandre Alahi.
\newblock Are socially-aware trajectory prediction models really
  socially-aware?
\newblock {\em Transportation research part C: emerging technologies},
  141:103705, 2022.

\bibitem{sadeghian2018trajnet}
Amir Sadeghian, Vineet Kosaraju, Agrim Gupta, Silvio Savarese, and Alexandre
  Alahi.
\newblock Trajnet: Towards a benchmark for human trajectory prediction.
\newblock {\em arXiv preprint}, 2018.

\bibitem{sadeghian2019sophie}
Amir Sadeghian, Vineet Kosaraju, Ali Sadeghian, Noriaki Hirose, Hamid
  Rezatofighi, and Silvio Savarese.
\newblock Sophie: An attentive gan for predicting paths compliant to social and
  physical constraints.
\newblock In {\em Proceedings of the IEEE/CVF conference on computer vision and
  pattern recognition}, pages 1349--1358, 2019.

\bibitem{salzmann2020trajectron++}
Tim Salzmann, Boris Ivanovic, Punarjay Chakravarty, and Marco Pavone.
\newblock Trajectron++: Dynamically-feasible trajectory forecasting with
  heterogeneous data.
\newblock In {\em Computer Vision--ECCV 2020: 16th European Conference,
  Glasgow, UK, August 23--28, 2020, Proceedings, Part XVIII 16}, pages
  683--700. Springer, 2020.

\bibitem{serra2018overcoming}
Joan Serra, Didac Suris, Marius Miron, and Alexandros Karatzoglou.
\newblock Overcoming catastrophic forgetting with hard attention to the task.
\newblock In {\em International Conference on Machine Learning}, pages
  4548--4557. PMLR, 2018.

\bibitem{shi2022motion}
Shaoshuai Shi, Li Jiang, Dengxin Dai, and Bernt Schiele.
\newblock Motion transformer with global intention localization and local
  movement refinement.
\newblock {\em arXiv preprint arXiv:2209.13508}, 2022.

\bibitem{silver2002task}
Daniel~L Silver and Robert~E Mercer.
\newblock The task rehearsal method of life-long learning: Overcoming
  impoverished data.
\newblock In {\em Conference of the Canadian Society for Computational Studies
  of Intelligence}, pages 90--101. Springer, 2002.

\bibitem{sun2022m2i}
Qiao Sun, Xin Huang, Junru Gu, Brian~C Williams, and Hang Zhao.
\newblock M2i: From factored marginal trajectory prediction to interactive
  prediction.
\newblock In {\em Proceedings of the IEEE/CVF Conference on Computer Vision and
  Pattern Recognition}, pages 6543--6552, 2022.

\bibitem{sun2020ernie}
Yu Sun, Shuohuan Wang, Yukun Li, Shikun Feng, Hao Tian, Hua Wu, and Haifeng
  Wang.
\newblock Ernie 2.0: A continual pre-training framework for language
  understanding.
\newblock In {\em Proceedings of the AAAI Conference on Artificial
  Intelligence}, volume~34, pages 8968--8975, 2020.

\bibitem{tong2022videomae}
Zhan Tong, Yibing Song, Jue Wang, and Limin Wang.
\newblock Videomae: Masked autoencoders are data-efficient learners for
  self-supervised video pre-training.
\newblock {\em arXiv preprint arXiv:2203.12602}, 2022.

\bibitem{varadarajan2022multipath++}
Balakrishnan Varadarajan, Ahmed Hefny, Avikalp Srivastava, Khaled~S Refaat,
  Nigamaa Nayakanti, Andre Cornman, Kan Chen, Bertrand Douillard, Chi~Pang Lam,
  Dragomir Anguelov, et~al.
\newblock Multipath++: Efficient information fusion and trajectory aggregation
  for behavior prediction.
\newblock In {\em 2022 International Conference on Robotics and Automation
  (ICRA)}, pages 7814--7821. IEEE, 2022.

\bibitem{vaswani2017attention}
Ashish Vaswani, Noam Shazeer, Niki Parmar, Jakob Uszkoreit, Llion Jones,
  Aidan~N Gomez, {\L}ukasz Kaiser, and Illia Polosukhin.
\newblock Attention is all you need.
\newblock {\em Advances in neural information processing systems}, 30, 2017.

\bibitem{vincent2008extracting}
Pascal Vincent, Hugo Larochelle, Yoshua Bengio, and Pierre-Antoine Manzagol.
\newblock Extracting and composing robust features with denoising autoencoders.
\newblock In {\em Proceedings of the 25th international conference on Machine
  learning}, pages 1096--1103, 2008.

\bibitem{wang2019learning}
Haohan Wang, Songwei Ge, Zachary Lipton, and Eric~P Xing.
\newblock Learning robust global representations by penalizing local predictive
  power.
\newblock {\em Advances in Neural Information Processing Systems}, 32, 2019.

\bibitem{wang2022ganet}
Mingkun Wang, Xinge Zhu, Changqian Yu, Wei Li, Yuexin Ma, Ruochun Jin,
  Xiaoguang Ren, Dongchun Ren, Mingxu Wang, and Wenjing Yang.
\newblock Ganet: Goal area network for motion forecasting.
\newblock {\em arXiv preprint arXiv:2209.09723}, 2022.

\bibitem{yang2019xlnet}
Zhilin Yang, Zihang Dai, Yiming Yang, Jaime Carbonell, Russ~R Salakhutdinov,
  and Quoc~V Le.
\newblock Xlnet: Generalized autoregressive pretraining for language
  understanding.
\newblock {\em Advances in neural information processing systems}, 32, 2019.

\bibitem{ye2021tpcn}
Maosheng Ye, Tongyi Cao, and Qifeng Chen.
\newblock Tpcn: Temporal point cloud networks for motion forecasting.
\newblock In {\em Proceedings of the IEEE/CVF Conference on Computer Vision and
  Pattern Recognition}, pages 11318--11327, 2021.

\bibitem{ye2022dcms}
Maosheng Ye, Jiamiao Xu, Xunnong Xu, Tongyi Cao, and Qifeng Chen.
\newblock Dcms: Motion forecasting with dual consistency and
  multi-pseudo-target supervision.
\newblock {\em arXiv preprint arXiv:2204.05859}, 2022.

\bibitem{yu2020spatio}
Cunjun Yu, Xiao Ma, Jiawei Ren, Haiyu Zhao, and Shuai Yi.
\newblock Spatio-temporal graph transformer networks for pedestrian trajectory
  prediction.
\newblock In {\em European Conference on Computer Vision}, pages 507--523.
  Springer, 2020.

\bibitem{yuan2021agentformer}
Ye Yuan, Xinshuo Weng, Yanglan Ou, and Kris~M Kitani.
\newblock Agentformer: Agent-aware transformers for socio-temporal multi-agent
  forecasting.
\newblock In {\em Proceedings of the IEEE/CVF International Conference on
  Computer Vision}, pages 9813--9823, 2021.

\bibitem{zeng2021lanercnn}
Wenyuan Zeng, Ming Liang, Renjie Liao, and Raquel Urtasun.
\newblock Lanercnn: Distributed representations for graph-centric motion
  forecasting.
\newblock In {\em 2021 IEEE/RSJ International Conference on Intelligent Robots
  and Systems (IROS)}, pages 532--539. IEEE, 2021.

\bibitem{zeno2018task}
Chen Zeno, Itay Golan, Elad Hoffer, and Daniel Soudry.
\newblock Task agnostic continual learning using online variational bayes.
\newblock {\em arXiv preprint arXiv:1803.10123}, 2018.

\bibitem{zhan2019interaction}
Wei Zhan, Liting Sun, Di Wang, Haojie Shi, Aubrey Clausse, Maximilian Naumann,
  Julius Kummerle, Hendrik Konigshof, Christoph Stiller, Arnaud de La~Fortelle,
  et~al.
\newblock Interaction dataset: An international, adversarial and cooperative
  motion dataset in interactive driving scenarios with semantic maps.
\newblock {\em arXiv preprint arXiv:1910.03088}, 2019.

\bibitem{zhang2022trajectory}
Lu Zhang, Peiliang Li, Jing Chen, and Shaojie Shen.
\newblock Trajectory prediction with graph-based dual-scale context fusion.
\newblock In {\em 2022 IEEE/RSJ International Conference on Intelligent Robots
  and Systems (IROS)}, pages 11374--11381. IEEE, 2022.

\bibitem{zhang2019sr}
Pu Zhang, Wanli Ouyang, Pengfei Zhang, Jianru Xue, and Nanning Zheng.
\newblock Sr-lstm: State refinement for lstm towards pedestrian trajectory
  prediction.
\newblock In {\em Proceedings of the IEEE/CVF Conference on Computer Vision and
  Pattern Recognition}, pages 12085--12094, 2019.

\bibitem{zhang2018overview}
Yu Zhang and Qiang Yang.
\newblock An overview of multi-task learning.
\newblock {\em National Science Review}, 5(1):30--43, 2018.

\bibitem{zhao2020tnt}
Hang Zhao, Jiyang Gao, Tian Lan, Chen Sun, Benjamin Sapp, Balakrishnan
  Varadarajan, Yue Shen, Yi Shen, Yuning Chai, Cordelia Schmid, et~al.
\newblock Tnt: Target-driven trajectory prediction.
\newblock {\em arXiv preprint arXiv:2008.08294}, 2020.

\bibitem{zhou2022hivt}
Zikang Zhou, Luyao Ye, Jianping Wang, Kui Wu, and Kejie Lu.
\newblock Hivt: Hierarchical vector transformer for multi-agent motion
  prediction.
\newblock In {\em Proceedings of the IEEE/CVF Conference on Computer Vision and
  Pattern Recognition}, pages 8823--8833, 2022.

\end{thebibliography}
}

\end{document}


\title{Traj-MAE: Masked Autoencoders for Trajectory Prediction}

\author{First Author\\
Institution1\\
Institution1 address\\
{\tt\small firstauthor@i1.org}
\and
Second Author\\
Institution2\\
First line of institution2 address\\
{\tt\small secondauthor@i2.org}
}
\maketitle

\section{Autobots Architecture.}
Autobots \cite{girgis2022latent} is a class of encoder-decoder architectures that process sequences of sets. This model is designed to process a tensor of dimensions $\emph{K} \times \emph{M} \times \emph{t}$ as input, where \emph{K} is the number of attributes of each agent, \emph{M} is the number of agents, and \emph{t} is the input sequence length. To transform the \emph{K}-dimensional vectors to a new space of dimension $\textit{d}_{K}$ (hidden size), it applies a row-wise feed-forward network (rFFN) to each row along the $\emph{t} \times \emph{M}$ plane. Following the addition of positional encoding (PE) to the \emph{t} axis, the encoder processes the tensor through \emph{L} layers of repeated multi-head attention blocks (MAB) that apply time encoding and social encoding to the time and agent axes, respectively. Finally, the encoder outputs the context tensor. In the decoder, the encoded map and the learnable seed parameters tensor are first concatenated and then passed through an rFFN. The resulting tensor is then processed through \emph{L} layers of repeated multi-head attention block decoder (MABD) along the time axis using the context from the encoder, followed by a MAB along the agent axis.
The output of the decoder is a tensor of dimensions $\textit{d}_{K} \times \emph{M} \times \emph{T} \times \emph{c}$, which can then be element-wise processed using a neural network to produce the desired output representation. \emph{T} is the output sequence length and \emph{c} is the number of modes.
AutoBot-Ego is a special case, which is similar to AutoBots but predicts future modes for only one agent in a scene.

\section{Implementation Details.}
For pre-training, we use an Adam optimizer \cite{kingma2014adam} with a fixed learning rate 1e-3. We set the same number of training stages as the number of pre-training strategies. When performing continuous pre-training, 
the number of steps for each strategy in all training steps adds up to $120k$.
In terms of the trajectory prediction task, we fine-tune Autobots on three challenging datasets, and utilize Adam as an optimizer. For the Argoverse dataset~\cite{chang2019argoverse}, we use Autobot-Ego as our baseline model, the initial learning rate is set at 3e-5, and for the Interaction~\cite{zhan2019interaction} and TrajNet++~\cite{sadeghian2018trajnet} datasets, Autobot is used as the baseline model, we set the initial learning rates at 5e-5 and 7e-5, respectively. We anneal the learning rate every $6k$ by a factor of 2 in the first $30k$ steps, and the total training steps are $120k$.
The batch size of training and testing of all the above tasks is 64. The Traj-MAE is implemented by PyTorch and all experiments can be done on a single V100.


\begin{table}[t]
\centering
\resizebox{\columnwidth}{!}{
\begin{tabular}{cccc} \toprule
Method & \hspace{-1.0cm}minADE & \hspace{-0.9cm}minFDE & \hspace{-1.1cm}MR \\ \hline
DESIRE~\cite{lee2017desire} & \hspace{-1.0cm}0.92 & \hspace{-0.9cm}1.77 & \hspace{-1.1cm}0.18 \\
MultiPath~\cite{chai2019multipath} & \hspace{-1.0cm}0.80 & \hspace{-0.9cm}1.68 & \hspace{-1.1cm}0.14 \\
TNT~\cite{zhao2020tnt} & \hspace{-1.0cm}0.73 & \hspace{-0.9cm}1.29 & \hspace{-1.1cm}0.09 \\
LaneRCNN~\cite{zeng2021lanercnn} & \hspace{-1.0cm}0.77 & \hspace{-0.9cm}1.19 & \hspace{-1.1cm}0.08 \\
TPCN~\cite{ye2021tpcn} & \hspace{-1.0cm}0.73 & \hspace{-0.9cm}1.15 & \hspace{-1.1cm}0.11 \\
mmTransformer~\cite{liu2021multimodal} & \hspace{-1.0cm}0.71 & \hspace{-0.9cm}1.15 & \hspace{-1.1cm}0.11 \\
DenseTNT~\cite{gu2021densetnt} & \hspace{-1.0cm}0.82 & \hspace{-0.9cm}1.37 & \hspace{-1.1cm}\textbf{0.07} \\
HiVT\cite{zhou2022hivt} & \hspace{-1.0cm}0.66 & \hspace{-0.9cm}0.96 & \hspace{-1.1cm}0.09 \\ 
DCMS\cite{ye2022dcms} & \hspace{-1.0cm}0.64 & \hspace{-0.9cm}\textbf{0.93} & \hspace{-1.1cm}- \\ 
GANet\cite{wang2022ganet} & \hspace{-1.0cm}0.67 & \hspace{-0.9cm}\textbf{0.93} & \hspace{-1.1cm}- \\ \hline
Autobot-Ego~\cite{girgis2022latent} & \hspace{-1.0cm}0.73 & \hspace{-0.9cm}1.10 & \hspace{-1.1cm}0.12 \\
\textbf{Traj-MAE} & \hspace{0cm}\textbf{0.60} \red{$\downarrow$ 18\%} & \hspace{-0.1cm}1.00 \red{$\downarrow$ 9\%} & \hspace{-0.1cm}0.09 \red{$\downarrow$ 25\%} \\ \toprule
\end{tabular}
}
\caption{\textbf{Comparison with state-of-the-art methods on the Argoverse validation set}.}
\label{table:argo_val}
\vspace{-5pt}
\end{table}

\begin{table*}[]
\centering
\begin{tabular}{c|c|ccc|ccc}
\hline
\multirow{2}{*}{Method} & \multirow{2}{*}{Pre-training strategy} & \multicolumn{3}{c|}{Training steps} & \multicolumn{3}{c}{Fine-tuning result} \\ \cline{3-8} 
 &  & Stage1 & Stage2 & Stage3 & minADE & minFDE & MR \\ \hline
\multirow{3}{*}{Continual Learning} & S & 120k & - & - & \multirow{3}{*}{0.664} & \multirow{3}{*}{1.075} & \multirow{3}{*}{0.108} \\
 & T & - & 120k & - &  &  &  \\
 & ST & - & - & 120k &  &  &  \\ \hline
\multirow{3}{*}{Multi-task Learning} & S & \multicolumn{3}{c|}{120k} & \multirow{3}{*}{0.693} & \multirow{3}{*}{1.089} & \multirow{3}{*}{0.113} \\
 & T & \multicolumn{3}{c|}{120k} &  &  &  \\
 & ST & \multicolumn{3}{c|}{120k} &  &  &  \\ \hline
\multirow{3}{*}{Continual Pre-training} & S & 60k & 30k & 30k & \multirow{3}{*}{\textbf{0.621}} & \multirow{3}{*}{\textbf{1.027}} & \multirow{3}{*}{\textbf{0.099}} \\
 & T & - & 90k & 30k &  &  &  \\
 & ST & - & - & 120k &  &  &  \\ \hline
\end{tabular}
\caption{\textbf{Continual Pre-training for trajectory reconstruction}. Note that 'S', 'T', 'ST' represent social masking, temporal masking, social and temporal masking strategy, respectively.}
\label{table:continual_trajectory}
\end{table*}

\begin{table*}[]
\centering
\begin{tabular}{c|c|ccc|ccc}
\hline
\multirow{2}{*}{Method} & \multirow{2}{*}{Pre-training strategy} & \multicolumn{3}{c|}{Training steps} & \multicolumn{3}{c}{Fine-tuning result} \\ \cline{3-8} 
 &  & Stage1 & Stage2 & Stage3 & minADE & minFDE & MR \\ \hline
\multirow{3}{*}{Continual Learning} & Po & 120k & - & - & \multirow{3}{*}{0.656} & \multirow{3}{*}{1.069} & \multirow{3}{*}{0.107} \\
 & Pa & - & 120k & - &  &  &  \\
 & B & - & - & 120k &  &  &  \\ \hline
\multirow{3}{*}{Multi-task Learning} & Po & \multicolumn{3}{c|}{120k} & \multirow{3}{*}{0.685} & \multirow{3}{*}{1.081} & \multirow{3}{*}{0.112} \\
 & Pa & \multicolumn{3}{c|}{120k} &  &  &  \\
 & B & \multicolumn{3}{c|}{120k} &  &  &  \\ \hline
\multirow{3}{*}{Continual Pre-training} & Po & 60k & 30k & 30k & \multirow{3}{*}{\textbf{0.627}} & \multirow{3}{*}{\textbf{1.033}} & \multirow{3}{*}{\textbf{0.102}} \\
 & Pa & - & 90k & 30k &  &  &  \\
 & B & - & - & 120k &  &  &  \\ \hline
\end{tabular}
\caption{\textbf{Continual Pre-training for map reconstruction}. Note that 'Po', 'Pa', 'B' represent point masking, patch masking, block masking strategy, respectively.}
\label{table:continual_map}
\vspace{-10pt}
\end{table*}

\section{Metrics.}
For the task to predict the ego-agent's future trajectory, we use minADE, minFDE and MR to evaluate our method, which are respectively the minimum Average Displacement Error, the minimum Final Displacement Error and the Miss Rate, respectively. Considering the multi-agents' future trajectories prediction task, we calculate ego-agent's prediction error and scene-level prediction error as defined by \cite{casas2020implicit} on TrajNet++. Similarly, MinJointADE, MinJointFDE and MinJointMR are used to calculate multi-agents' prediction error. CrossCollisionRate represents the frequency of collisions happening among the predictions of all agents and EgoCollisionRate represents the collisions happening between ego-agent and others. When only considering those modalities without cross collision, Consistent MinJointMR is calculated as the case's miss rate.
All of the above metrics are the lower the better.
The lower metrics reflect better performance.

\section{Experimental Results}
\noindent\textbf{Argoverse validation set.} In Table \ref{table:argo_val}, we verify our method on the Argoverse validation set and demonstrate superior performance, achieving the lowest minADE score of 0.60 compared to other approaches. Furthermore, in comparison to our baseline model, our Traj-MAE exhibits a noteworthy reduction in minFDE and MR from 1.10 to 1.00 (9\%) and 0.12 to 0.09 (25\%), respectively.


\begin{table}[t]
\centering
\resizebox{\linewidth}{!}{%
\begin{tabular}{ccccc}
\toprule
Num & Modules & minADE & minFDE & MR \\ \hline
E1 & baseline & 0.730 & 1.100  & 0.120 \\
E2 & E1 + map encoder & 0.732  & 1.096 & 0.119 \\
E3 & E2 + trajectory pre-train & 0.621  & 1.027 & 0.099 \\
E4 & E3 + map pre-train & \textbf{0.604} & \textbf{1.003} & \textbf{0.092} \\ \toprule
\end{tabular}
}
\caption{\textbf{Ablation on the effectiveness of our modules}.}
\label{table:ablation_component}
\vspace{-8pt}
\end{table}

\noindent\textbf{Continue Pre-training.}
To investigate the properties of our proposed continual pre-training, we compare it with continual learning and multi-task learning.
As we can see from Table \ref{table:continual_trajectory} and Table \ref{table:continual_map}, all method train each strategy with the same steps for a fair comparison.
Our proposed continual pre-training achieves the best fine-tuning results.
Moreover, multi-task learning, which learns multiple strategies at the same time, achieves little improvement compared to our baseline model. This could be that reconstructing with so many strategies simultaneously is too hard for the model to learn, thus even hurting the model's representation ability. We hope that future work will explore the different novel designs of continual learning and multi-task learning to make them suitable for trajectory prediction.

\noindent\textbf{Module Analysis.}
Table \ref{table:ablation_component} shows the effectiveness of different modules of our Traj-MAE.
First, we find that adding a map encoder directly to process the HD map brings little improvement to Autobots, and even hurts the model's performance on minADE. 
Pre-training the trajectory encoder can improve the accuracy significantly, especially on minADE.
What's more, the accuracy can be further improved when we pre-train the map encoder on the basis of $E3$.
Thus, the validity of our proposed trajectory pre-training module and map pre-training module can be proved.

\section{Visualization}
\begin{figure*}
    \centering
        \includegraphics[width=\linewidth]{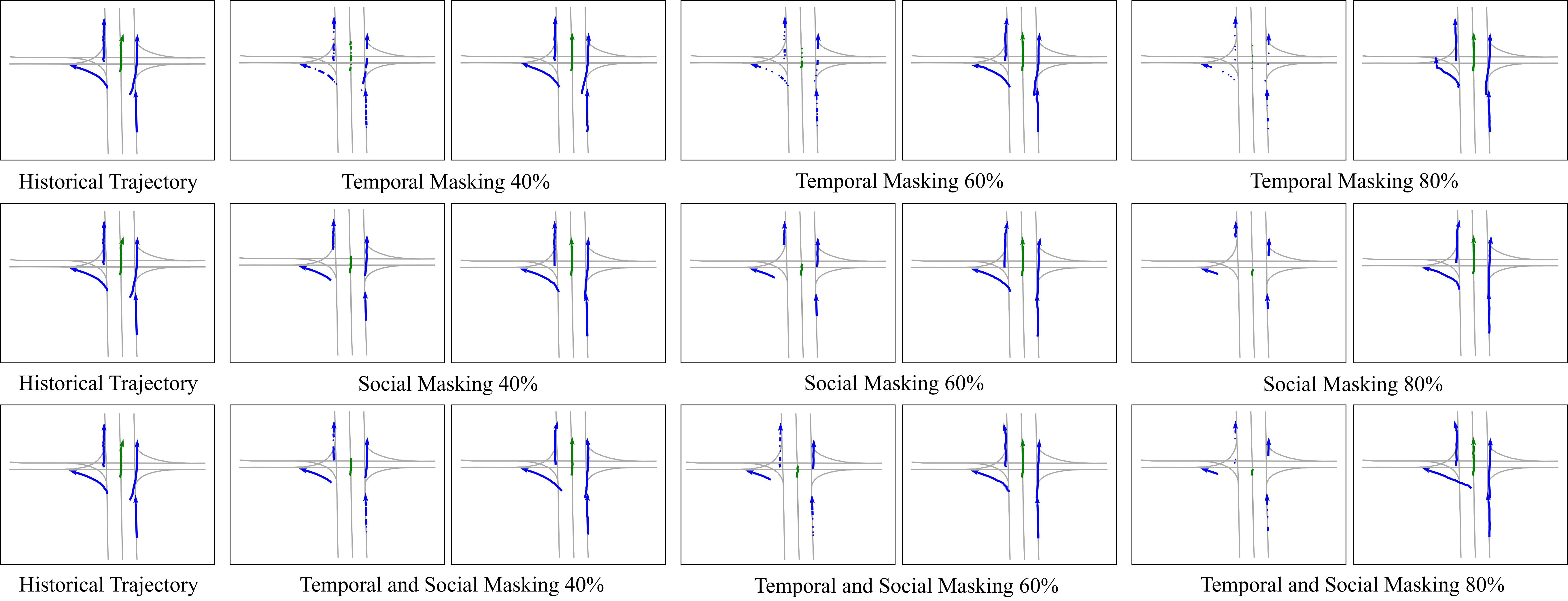}
    \caption{\textbf{Reconstruction results on historical trajectory.}}
    \label{fig:sup3}
\end{figure*}

\begin{figure*}
    \centering
        \includegraphics[width=\linewidth]{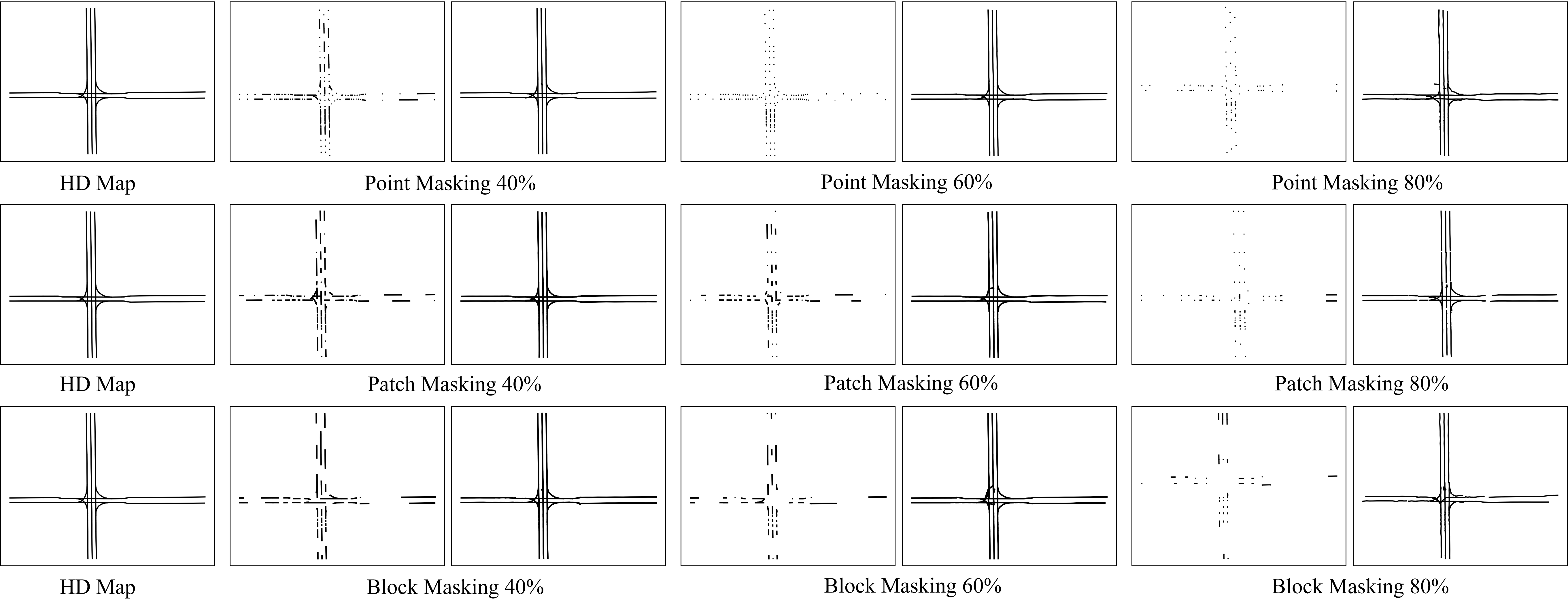}
    \caption{\textbf{Reconstruction results on HD map.}}
    \label{fig:sup4}
\end{figure*}

\begin{figure*}[ht]
    \centering
        \includegraphics[width=\linewidth]{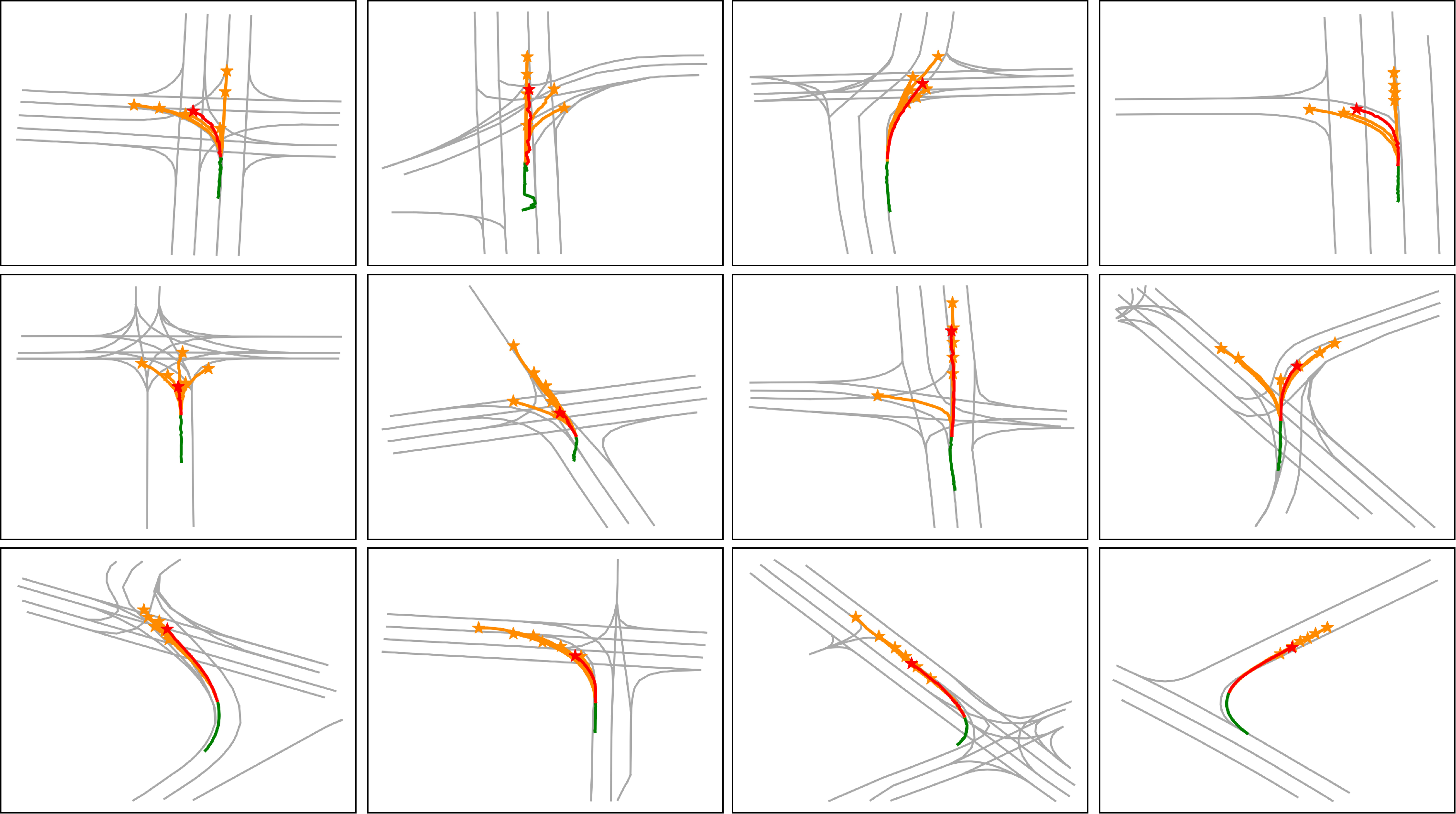}
    \caption{\textbf{The motion forecasting results on the Argoverse validation set.} The historical trajectory of the target agent is in green, predicted trajectories in orange and ground truth in red, respectively.}
    \label{fig:sup5}
\end{figure*}

We show several examples of reconstruction in Figure \ref{fig:sup3} and Figure \ref{fig:sup4}. 
The motion forecasting results on the Argoverse validation set are shown in Figure \ref{fig:sup5}.
Samples are all chosen from the Argoverse validation set.

\clearpage

\begin{figure*}
    \clearpage
    \centering
        \includegraphics[width=\linewidth]{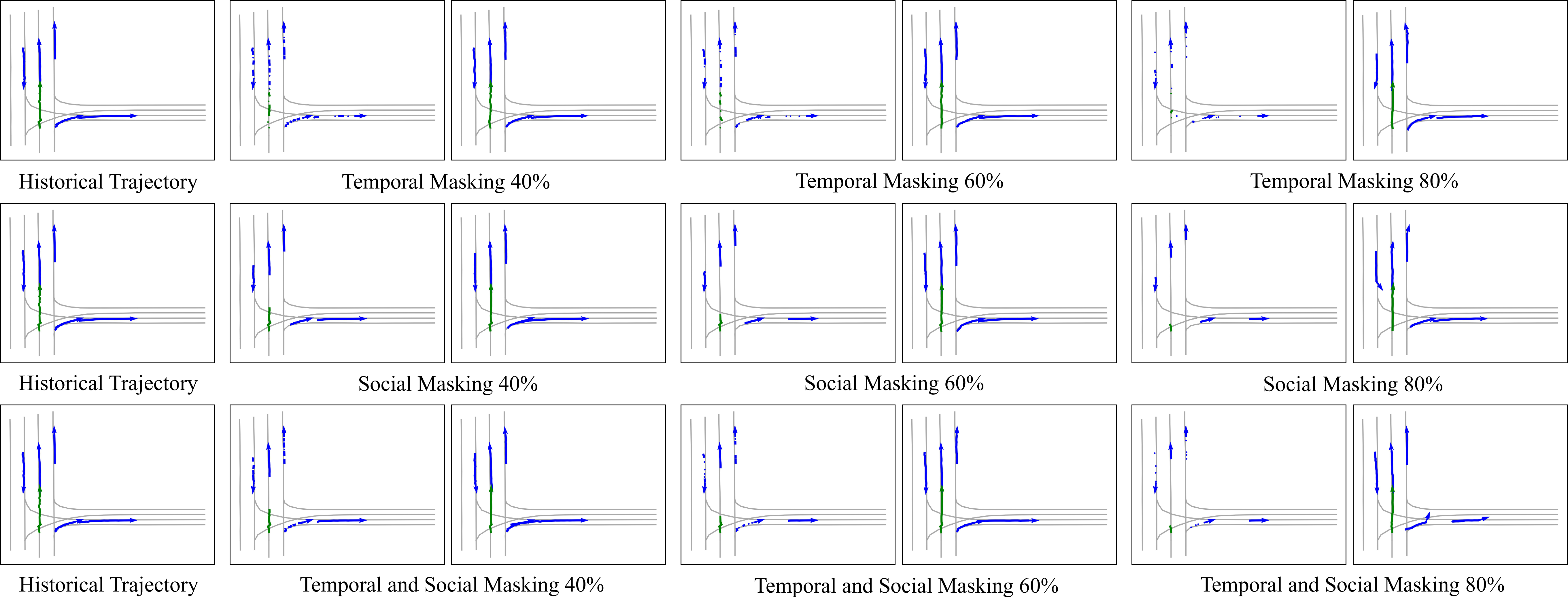}
    \label{fig:sup1}
\end{figure*}

\begin{figure*}
    \centering
        \includegraphics[width=\linewidth]{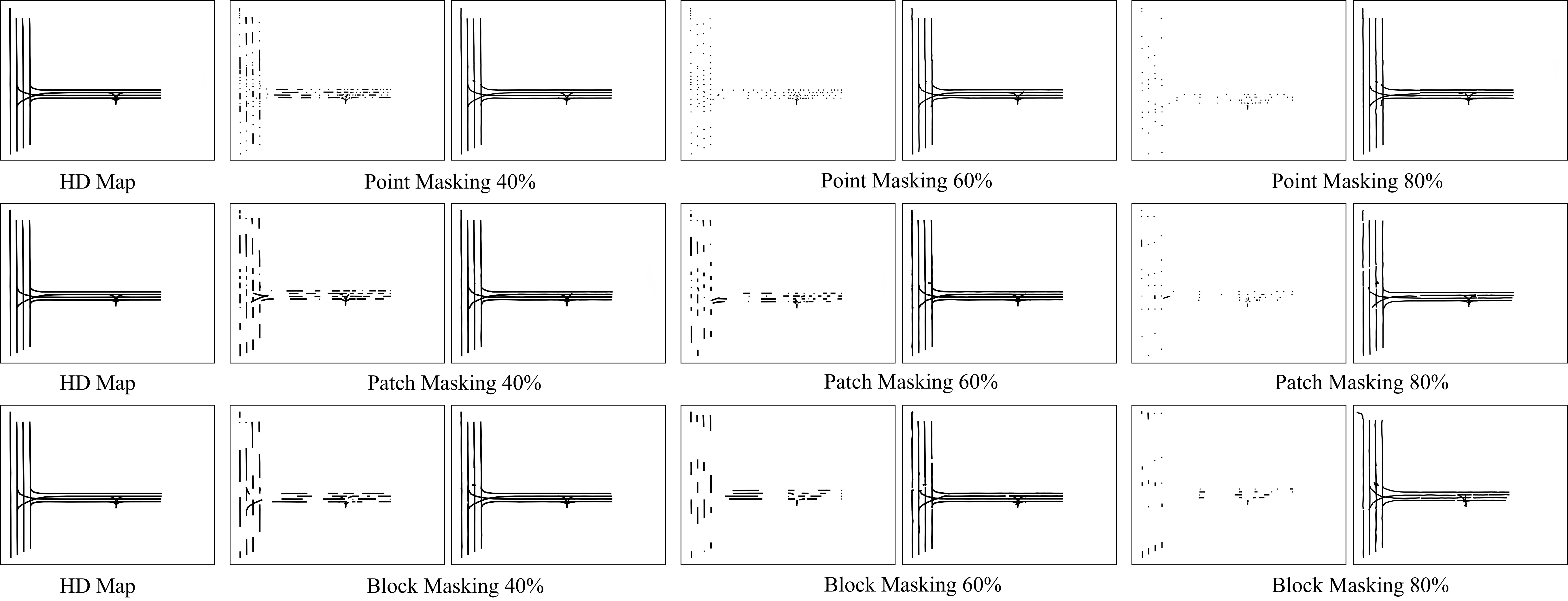}
    \label{fig:sup2}
\end{figure*}
\clearpage

\clearpage

{\small
\bibliographystyle{ieee_fullname}
\bibliography{egbib}
}